\pdfoutput=1

\documentclass[11pt]{article}

\usepackage{ACL2023}

\usepackage{times}
\usepackage{latexsym}

\usepackage[T1]{fontenc}

\usepackage[utf8]{inputenc}

\usepackage{microtype}

\usepackage{inconsolata}

\usepackage{xspace}

\newcommand{\riskharm}{\textsc{risk of harm}\xspace}
\newcommand{\riskfail}{\textsc{risk of failure}\xspace}
\newcommand{\spec}{\textsc{expertise}\xspace}

\definecolor{c_purple}{HTML}{4F2EB9}
\definecolor{c_orange}{HTML}{B14C0C}
\definecolor{c_blue}{HTML}{0095B1}
\definecolor{c_green}{HTML}{7A904C}
\definecolor{c_pink}{HTML}{A26088}

\usepackage{graphicx}
\usepackage{booktabs}
\usepackage{rotating}
\usepackage{subfig}
\usepackage{amsmath}
\usepackage{array}
\newcolumntype{R}[1]{>{\raggedleft\arraybackslash}p{#1}}
\newcolumntype{L}[1]{>{\raggedright\arraybackslash}p{#1}}
\usepackage{subcaption}

\title{Risks and NLP Design:  A Case Study on Procedural Document QA}

\author{Nikita Haduong$^1$ \quad Alice Gao$^1$ \quad  Noah A.~Smith$^{1,2}$ \\
$^1$Paul G.~Allen School of Computer Science \& Engineering, University of Washington \\
$^2$Allen Institute for Artificial Intelligence \\
\texttt{\{qu,atgao,nasmith\}@cs.washington.edu} }

\begin{document}
\maketitle

\begin{abstract}

    As NLP systems are increasingly deployed at scale, concerns about their potential negative impacts have attracted the attention of the research community, yet discussions of risk have mostly been at an abstract level and focused on generic AI or NLP applications.  We argue that clearer assessments of risks and harms to users---and concrete strategies to mitigate them---will be possible when we specialize the analysis to more concrete applications and their plausible users.  As an illustration, this paper is grounded in cooking recipe procedural document question answering (ProcDocQA), where there are well-defined risks to users such as injuries or allergic reactions.  Our case study shows that an existing language model, applied in ``zero-shot'' mode, quantitatively answers real-world questions about recipes as well or better than the humans who have answered the questions on the web.  Using a novel questionnaire informed by theoretical work on AI risk, we conduct a risk-oriented error analysis that could then inform the design of a future system to be deployed with lower risk of harm and better performance.  
    
    \end{abstract}

    \section{Introduction}
    
    Much of the current discussion about AI---in both the research community and the broader public---focuses on the tension between deployment of systems whose behavior is nearly indistinguishable from humans (\citealp{Clark2021Allt}, \emph{inter alia}) and understanding the potential consequences of such deployment, including fairness, reliability, and other social and ethical implications (\citealp{Tan2021ReliabilityTF,Jacobs2021MeasurementAF,privacy_democracy,berkman_humanrights}, \emph{inter alia}).  A common theme is the lack of rigorous assessment or guidelines for deploying models to end users \citep{Tan2022TheRO,predictability_surprise_lm}, with work in mitigating harms operating broadly over large, diverse settings (\citealp{blodgett-etal-2020-language, buiten_2019, ZHANG2022113800, bender-friedman-2018-data}).

            \label{sec:attributes}
        \begin{figure}[ht]
        \centering
        \includegraphics[width=0.43\textwidth]{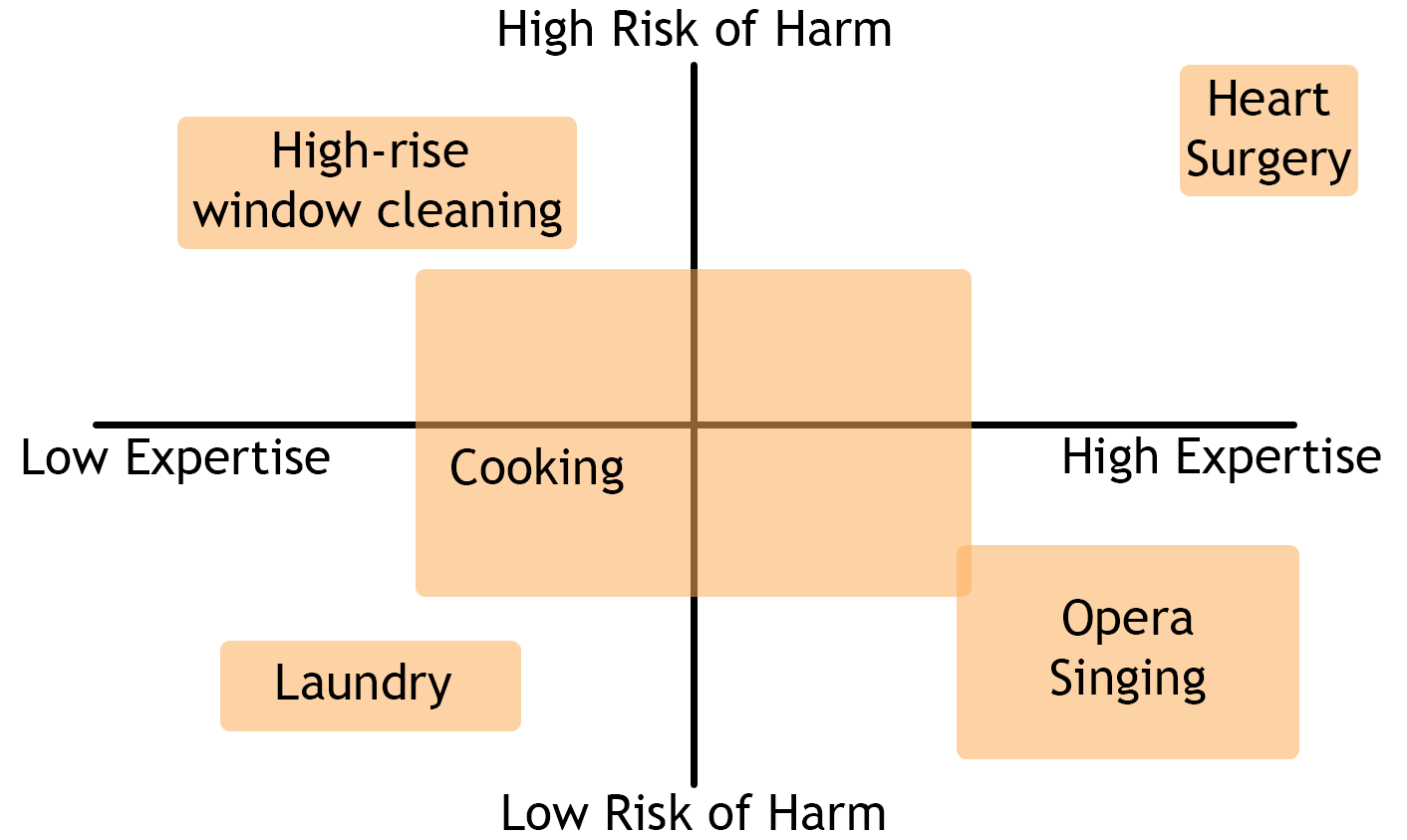}
        \caption{Dimensions characterizing procedural documents that can assist with estimating potential harms:  \riskharm\ to the user or environment and the \spec\ required for the user to successfully complete the procedure.}
        
        \label{fig:risk_specialization}
    \end{figure}

    We propose specializing the analysis of potential harms to more concrete applications, with identifiable user communities, to help close the gap between theoretical research on harms and research on real systems and users.
    We focus on question answering grounded in procedural documents (ProcDocQA), instructions written for a user to follow in order to complete a real-world task without supervision or assistance (e.g., cooking recipes), with discrete states of success. ProcDocQA can further be characterized along dimensions of \riskharm\ to the user and environment and the \spec required for a user to complete the procedure successfully. By articulating clear user goals (i.e., executing an instruction successfully), designers can more systematically assess the interplay of risks and system performance. 
    
    We introduce the first version of a Risk-Aware Design Questionnaire (RADQ) providing questions whose answers will be actionable for NLP designers of user-facing applications and conduct a case study in cooking recipes to illustrate how system designs evolve with the discovery of new risks. The case study shows how, despite zero-shot GPT-3 text-davinci-003 \citep{gpt3} achieving performance that is quantitatively on par with human-written answers, a deeper inspection of multiple answers per question reveals errors that will require application-specific resolutions. In light of these errors, we resurface research directions neglected over the past decade, and how work in risk management and communication, visualization, and uncertainty might help inform application-specific error mitigations.

    \section{ProcDocQA}

    Question answering is a mature NLP task with a diverse set of research datasets spanning many text and information domains, but risks and harms of question answering systems are underexplored, with work primarily in open-domain web question answering \citep{risk_webqa,Dhuliawala2022CalibrationOM}, user experience of a system (e.g., \citealp{Wang2021ControllingTR}), or privacy and security of users in an information retrieval stage of the system (e.g., \citealp{Wu2021ABF}). We refine the QA task to \emph{procedural} documents, which we argue enables more meaningful assessment of risks while maintaining a relatively high level of abstraction and large prospective user populations. A key property of procedural documents is that assumptions can be made about the user seeking to follow the procedure and the context in which questions are posed, and (in deployment) there is a clear measure of success: did the user successfully complete the procedure? 
    
    Assumptions about the user allow us to characterize genres and procedures within ProcDocQA along dimensions of \riskharm\ to the user and environment, concrete harms to specific entities that are more easily conceptualized than broad abstract harms to populations or society (as in \citealp{Tan2022TheRO}, \citealp{Lee2020InnovatingWC}, and \citealp{mental_health_biases}), and \spec, skill  required to successfully complete a procedure (Figure \ref{fig:risk_specialization}). For instance, the \riskharm\ of performing heart surgery can result in the death of the patient, and the surgeon requires high \spec\ to perform the operation. Doing laundry has a range in \spec\ due to knowledge required to launder a variety of fabrics (e.g., jeans vs.~a suit jacket), but there is low \riskharm\ (e.g., damaged clothing). For every instruction and task, there is an additional \riskfail, where the user may fail to successfully complete the instruction (which may also lead to \riskharm).
    We can now analyze how outputs of a ProcDocQA system affect \riskfail\ and \riskharm\ if the system is not calibrated toward the appropriate \spec\ of users. Note that \riskharm, \spec, and \riskfail\ can apply to every granularity of ProcDocQA: the overall genre (e.g., cooking), specific tasks (e.g. baking cookies), and individual instructions  (e.g., chop onions).

    \subsection*{Risk-Aware Design Questionnaire}

\begin{table*}[t]
\centering \small
\begin{tabular}{p{0.55\linewidth} | p{0.38\linewidth}}
\toprule
Question                                                                                                                                                                                                                                                                                                                                     & Purpose                                                                                                                                                                                                                                                                                                    \\ 
\midrule
Q1.1. Who are the \textbf{users} of the procedural document and what are the prerequisites for a user to be able to complete the procedure successfully? \newline Q1.2 What \textbf{tools} and \textbf{materials} are required for the task, and what are potential harms to the agent or environment if tools and materials are handled incorrectly? & To understand the demographics, values, and knowledge of the users to make appropriate assumptions when modifying system output (D in the DOCTOR framework; \citealp{Tan2021ReliabilityTF}). Grounds \riskharm\ and \spec\ in specific tasks/users.  \\ 
\hline
Q2. What are the most common \textbf{error types }present in outputs, and for each error type, what are its \textbf{potential harms}? In what contexts (question/answer types) do the error types appear? With respect to Q1, are some errors \textit{desirable}?                                                                            & To discover model output instability, revealing hidden potential for \riskharm, and inform designs for mitigations against such harms to lower \riskfail.                                                                                                                     \\ 
\hline
Q3.1 What are the upper and lower limits of \textbf{vagueness} in natural language responses to be effective? What are the effects of answers that are too vague, or too precise? \newline Q3.2 How much \textbf{confidence} should or can be expressed in the response?                                                                             & To calibrate system output to match user values and \spec\ (from Q1.1), thus improving user experience and lowering \riskfail.                                                                                                                                                             \\ 
\hline
Q4. When should the model \textbf{decline} to answer? What are the potential effects of returning \textbf{incorrect} answers? & To avoid returning low quality or incorrect answers that increase \riskfail or negatively impact user experience. \\ 
\hline
Q5. How should \textbf{multiple possible answers} be combined or reconciled before presenting a final response (e.g., a list of possible answers) to the user, and what are potential \textbf{consequences of confusion} for different reconciliation designs?                                                                               & To determine appropriate final responses to present to the user and in what manner, which can improve system helpfulness, thus lowering \riskfail.                                                           \\ 
\hline
Q6. What are possible harms that can arise from \textbf{user error/interpretation} of a response?                                                                                                                                                                                                                                            & To design preventative measures for inevitable human errors, reducing \riskfail.                                                                                                                   \\
\bottomrule
\end{tabular}
\caption{Risk-Aware Design Questionnaire for ProcDocQA.  Easily adaptable towards other user-facing applications.}\label{sec:worksheet} \label{tab:radq}
\end{table*}

The \riskharm and \spec levels illustrate, at a high level, how different end-user scenarios might affect QA system design, namely a system working with high \riskharm tasks may want to require high confidence answers verifiable by retrieved sources. Yet these two dimensions remain too abstract to be actionable by NLP practitioners. Therefore, in Table~\ref{tab:radq}, we propose the first version of a more detailed Risk-Aware Design Questionnaire (RADQ) to guide the design of a ProcDocQA system. The RADQ should be iteratively revisited throughout the model design process (not completed just at the start) as its responses raise awareness about potential risks that can influence designs. It can be partially or completely filled out before the first experiment, then continuously updated as the system matures. Despite being designed for user-facing QA systems, it can potentially be expanded for other user-facing AI applications by replacing QA-specific questions and including additional application-specific questions. For example, in a restaurant recommendation system, we might remove Q1.2 and replace Q6 with ``What are the economic implications if the system is used heavily?''

    \section{Case Study: ProcDocQA for Recipes}
    \label{sec:casestudy}

    We present a case study on cooking recipes, a genre of procedural documents with tasks that span a large range of \riskharm\ and \spec\ required for its tasks, but narrow the scope to homestyle recipes, which require less \spec\ and have lower \riskharm\ than professional-style recipes. 
    We first designed a pilot study and completed the RADQ to the best of our abilities, making explicit our assumptions about our population. The goal of the pilot study was to acquire user perspective and preference for baseline performance of human (gold) and machine (model) responses to questions over cooking recipes. Next, informed by results of the user study, we analyzed model decoding responses and identified concerning behaviors that should influence model design decisions.  In \S\ref{sec:radq-redux}, we return to the RADQ and propose ways in which the model design could be updated to be more \riskharm\ and \riskfail\ aware.

     \subsection{RADQ Initial Completion}\label{sec:radq-init}

    We describe how completing the RADQ to the best of our abilities contributed to user study questions and designs. After the user study, we update our RADQ responses in \S\ref{sec:rad-poststudy}, informed by research questions in multidisciplinary work.
    
    \paragraph{Q1} Users are home cooks who range in experience from novice to advanced. Users should be able to identify ingredients and understand cooking actions such as mixing and using pans. Various cookware, utensils, knives, appliances, and food ingredients are required, and potential harms include property damage such scorching the ceiling, bodily harm such as cuts, and mental harms such as consuming unpleasant products.
    
    \paragraph{Q2} While testing models, we observed infrequent undesirable behavior that led us to believe straightforward model use was not ready for deployment, motivating our study. 
    Consider the following:

    \textbf{Question:} Where do I go to buy a grit cooking mix in beijing China?  
    
    \textbf{GPT-3:} \textbf{I} bought a mix from \textbf{Trader Joes}, they have a great selection,and they even have a mix that is made with a mix of smoked gouda cheese and garlic powder.
    
    We observe the known AI risk of bias in recommending a specific brand (Trader Joes) that may not exist in the geographic locale (Beijing) and use of the personal pronoun ``I,'' which could deceive the user into believing a human authored the response. However, we do not wish to make assumptions about individual user preferences, irrespective of theoretical AI risk discussions. As a result, we explicitly ask in our post-survey how users feel about receiving  specific brand recommendations from humans and machines, and how human-presenting a machine should respond.
    
    \paragraph{Q3} We observe frequent hedging in human responses (e.g., ``\textit{I think} it would be too sweet but you can try it'') and expect such responses to provide lower satisfaction or helpfulness to a user. Answers that are too vague or precise may increase \riskfail\ because users are unable to effectively use such responses. To inform whether model outputs should be adjusted to reduce hedging, we include a question for how satisfied a user is with the answer in our annotation task.
    
    \paragraph{Q4} Questions relating to health that can carry high \riskharm\ such as asking for substitutes due to dietary restrictions should not be answered unless the model can retrieve sources, as distributing incorrect health information can have severe consequences resulting in hospitalizations and deaths. Incorrect answers for low \spec\ and low \riskharm\ tasks such as making pancakes can result in a failed final product, but the highest \riskharm\ to the user, burns, stems from incorrectly operating necessary appliances, rather than the procedure failure.
    
    \paragraph{Q5} Human responses often suggest multiple possible reasons for failure or ingredient substitutions in a text list, so we assume a similar format. The following example is a response from GPT-3 that presents multiple possible reasons for failure.

    \textbf{Paraphrased Question:} The madeleines do not stay fresh after a few hours and are very dry. I tried freezing them, but they are worse. Tips? \newline
    \textbf{GPT-3:}  [...] \textbf{First}, make sure you are using fresh, high-quality ingredients. \textbf{Second}, be sure to measure your ingredients carefully. \textbf{Third}, be sure to bake the blondies long enough - they should be golden brown and set in the center when they are done. \textbf{Finally}, be sure to let the blondies cool completely before cutting or serving. I hope this helps!

    \paragraph{Q6} User errors such as overkneading bread dough can result in disappointment with the resulting unsuccessful baked good. Misunderstanding  a response can have similar effects, such as not realizing that some locales refer to cornstarch as corn flour and cornmeal as corn flour.

    \subsection{User Perspective Study}
    
    We use GPT-3 text-davinci-003 with default parameters\footnote{Temperature $= 0.7$, $p=1$, access dates in  8--11/2022.} in a zero-shot setting to generate answers for questions from a custom dataset of blog  recipes collected from CommonCrawl (Appendix \ref{app:uqa--details}). The GPT-3 prompt was a concatenation of ingredients, instructions, the question, and ``Answer:'' (example prompts available in Appendix \ref{app:uqa--details} Table \ref{tab:gpt3PromptsContext}).

    Manual inspection of GPT-3 outputs revealed few NLG errors as described in the Scarecrow error analysis framework \citep{dou2021scarecrow}. Rather than create a recipe-specific extension of Scarecrow, we developed an annotation scheme for how responses could be improved along  improvement categories of concision, verbosity, and miscellaneous (Appendix \ref{app:categoriesImprovement}). Items within improvement categories were cooking-specific (e.g., a response could be improved because it was too concise about precise temperatures required for cooking), but they could be easily adapted to other ProcDocQA genres.

    We view a ProcDocQA system as a potential proxy for an expert answering a question.  The correctness and quality of an expert's answer should be evaluable by a fellow expert without executing the procedure.  Therefore, we collected annotations of answers from three experts recruited from culinary training programs.  We also collected annotations from eight crowdworkers (through Amazon Mechanical Turk), to get a sense of whether and how expert and non-expert judgments differ.\footnote{The study was exempted by our institution's IRB.} All annotators were located in the USA.
    
 Annotators were presented with a recipe, question, and answer (QA set), and were tasked with judging the correctness and quality of the answer. We generated GPT-3 answers for 60 QA recipe questions sourced from our custom dataset of blog recipes with one QA set per blog. Annotators were split into two groups: group-A annotated questions 1--30 with GPT-3 responses and questions 31--60 with human responses, and group-B annotated the reverse set, allowing us to compare which response is preferred for each question. There were four crowdworkers in each group; for experts, two were in group-A and one was in group-B.
 
 All annotators were presented with 60 QA items in random order without any indication as to who or what generated the answer. Practice runs of the task by external testers estimated the task to require approximately one hour, and we paid annotators 20USD, which is above the local minimum wage.\footnote{Crowdworkers spent 1--4 hours on the task with a median duration of 2 hours, and experts were ensured a pay rate of 20USD per hour.} The most common type of question asked was about ingredient substitutions, followed by ingredient and instruction clarification (Appendix Figure  \ref{fig:questionType}). The task also included a pre- and post-survey requesting information about demographics and user preferences regarding cooking question answering (Appendix \ref{app:caseStudy}).

    \subsection{Results}
    
    \begin{figure}
        \centering
        \includegraphics[width=.48\textwidth]{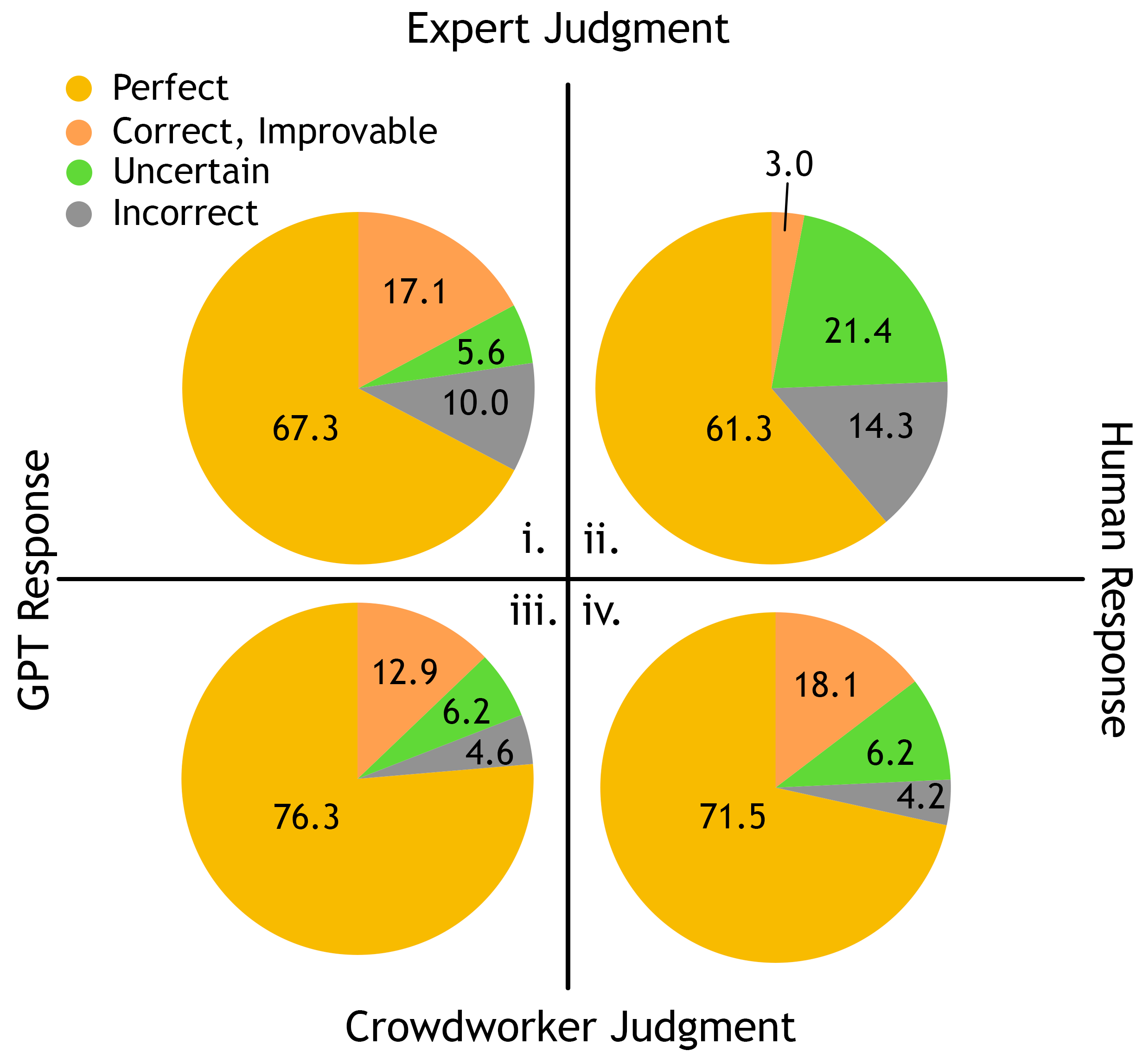}
        \caption{Annotators judged answers for correctness and could state their uncertainty about the answer correctness. Correct answers were judged for how they could be improved. Perfect answers required no change. Responses in i.~and ii.~were judged by experts, and iii.~and iv.~had crowdworker judges. GPT-3 generated responses in i.~and iii. Human-written answers were judged in ii.~and iv. Inter-annotator agreement about answer correctness was low for each group (Krippendorf's $\alpha<.5$), suggesting \spec and experience influence the perception of a correct answer. }
        
        \label{fig:caseResults}
    \end{figure}

    Overall, GPT-3 had strong performance, performing similarly to the human baseline, as judged by both crowdworkers and experts (Figure \ref{fig:caseResults}). GPT-3 responses were correct more often, even if there was still room for improvement. Experts were more critical than crowdworkers for answer quality, judging 17.1\% of GPT-3 responses correct but improvable vs.~12.9\% by crowdworkers. Crowdworkers gave 94.4\% of GPT-3 responses the highest satisfaction rating on a Likert scale from 1--5 as compared to 90.3\% of human responses, and experts gave 53.8\% of GPT-3 responses a satisfaction rating of 5 as compared to 40.0\% of human responses. Both GPT-3 and human responses were generally considered too concise: 52--55\% of correct answers annotated by experts had room for improvement in the concise category, and 80--85\% of crowdworker responses had room for improvement in the concise category (Appendix \ref{app:improvementResults}). Example annotation responses can be found in Appendix \ref{sec:exampleAnnotations}. Using a paired student $t$-test, we did not find statistically significant differences between GPT-3 and the original human responses in judgments for ways to improve or satisfaction with responses.

    \begin{table}[htb]
\centering
\begin{tabular}{lr} 
\toprule
 Behavior                   & \%          \\ 
\midrule
Output instability                     & 75.0    \\
Recommendations                 & 1.7 \\
Leading question agreement          &  5.0      \\
Hallucination & 18.3    \\
Language style     &  43.3  \\
Scarecrow \citep{dou2021scarecrow} errors    &  16.7    \\
Doesn't answer question         &  1.0   \\
\midrule
Perfect (no unexpected behavior) &  13.3  \\
\bottomrule
\end{tabular}
\caption{Percentage of prompts for which each behavior was present in at least one of the ten responses generated.  $N=60$. Multiple behaviors could be present in each prompt. }\label{tab:decodingErrorStats}
\end{table}
    
    \subsection{MultiDecoding Analysis} \label{sec:decoding_analysis}
 
Low error rates in GPT-3 responses, as rated by human annotators, imply that we only have a small sample of errors for analyzing potential harmful impacts.  Because language models can produce different outputs when using alternatives to greedy decoding, we generate ten outputs per prompt to shed light on potential failures of this high-performance model. When comparing the outputs to each other, the first author discovered several frequent classes of errors: output instability, recommendations, leading question agreement, hallucination, and language style, in addition to the Scarecrow errors ``needs Google,'' ``off prompt,''  ``self-contradiction,'' and outputs that do not answer the question.

We discuss each error type through the lenses of \riskharm\ to the user and environment, the \spec\  of the user, and \riskfail\ to complete the procedure. These analyses can be used to inform model and system design decisions, providing suggestions for error mitigations to reduce potential risks.  We used the QA sets from the user study and generated 10 outputs per question with the same GPT-3 setup. Only 13.3\% of prompts had 10 error-free outputs (Table \ref{tab:decodingErrorStats}).

    \begin{figure}
        \centering
        \includegraphics[width=.45\textwidth]{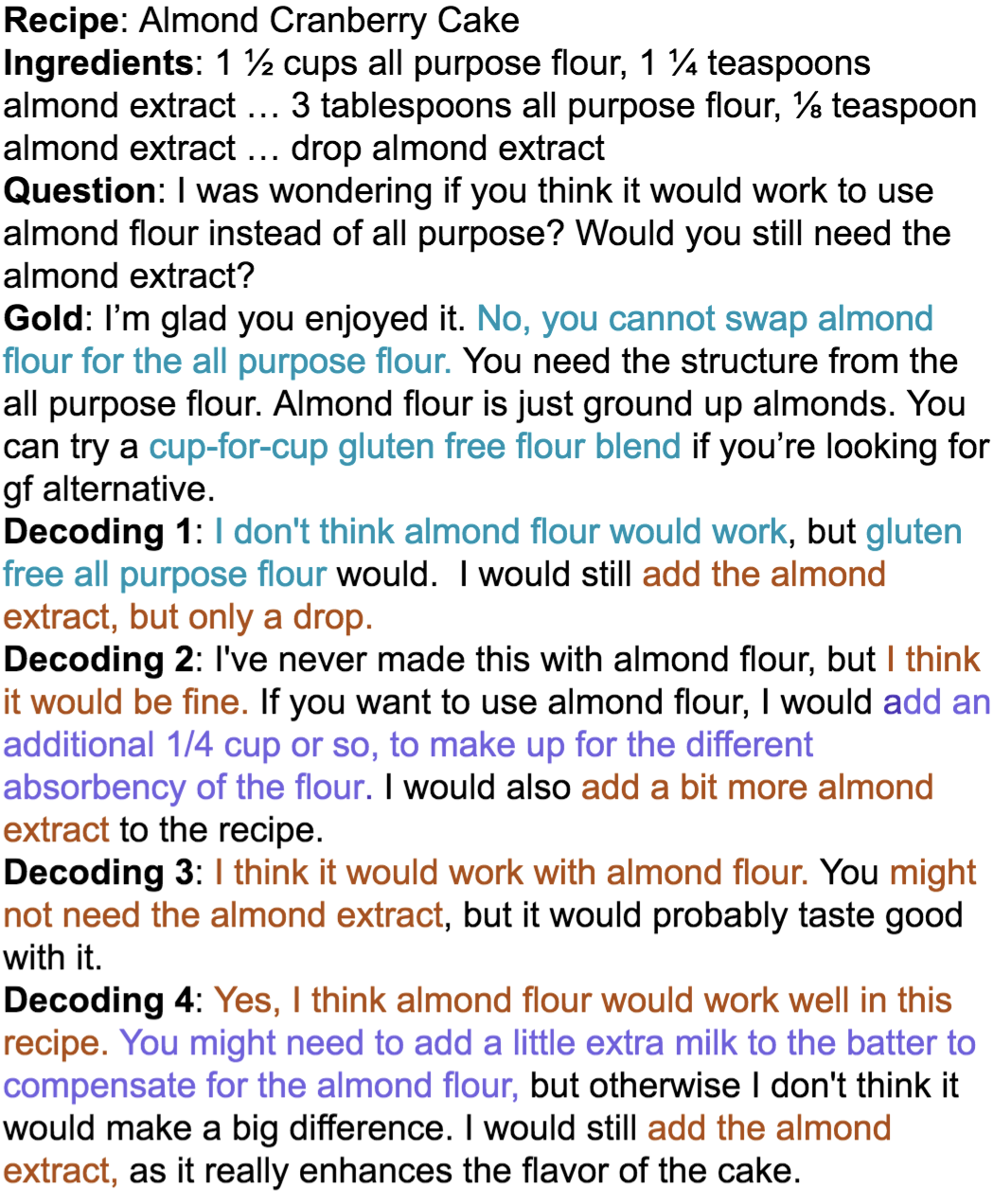}
        \caption{\textbf{Output instability}. The \textcolor{c_blue}{reference answer} states that you cannot swap the almond flour for all purpose flour. Decoding 1 agrees,  while decodings 2--4 state the \textcolor{c_orange}{opposite}. All decodings suggest different usage of \textcolor{c_orange}{almond extract}. Decodings 2 and 4 also suggest \textcolor{c_purple}{contrasting information} regarding the absorbancy of almond flour.}
        
        \label{ex:almondFlour}
    \end{figure} 
    
\begin{figure}
    \centering
    \includegraphics[width=.45\textwidth]{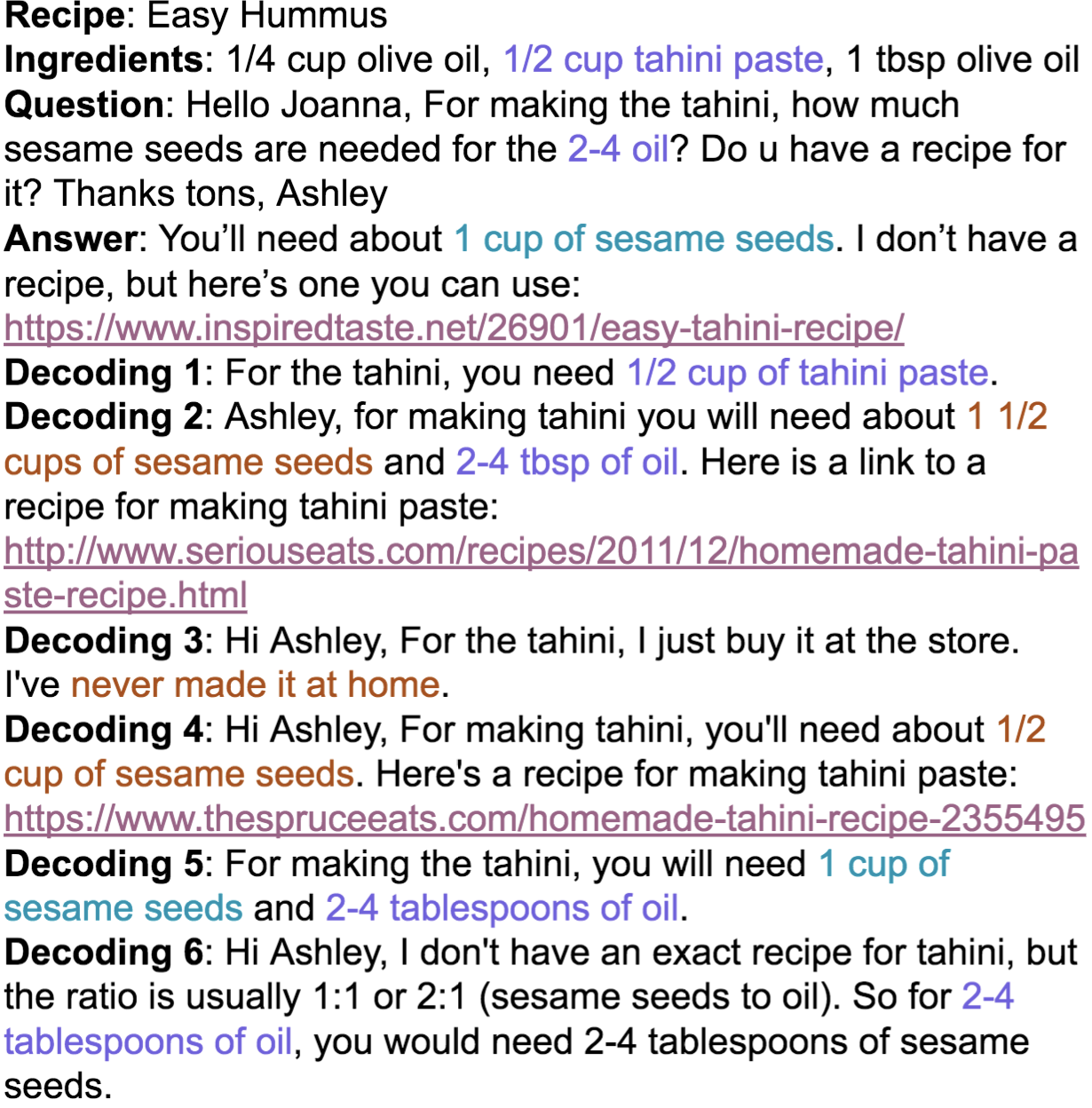}
    \caption{\textbf{Leading question agreement, hallucination, and recommendations}. The question includes \textcolor{c_purple}{contextual information} ``2-4 oil'' which decodings 2 and 5 use within their responses. Decodings 1 and 4 appear to use the \textcolor{c_purple}{1/2 cup} contextual information from the ingredients list rather than answer the question. Decodings 2 and 4 \textcolor{c_pink}{recommend} different recipe URLs that do not exist.
 }
    \label{ex:tahini}
\end{figure}

\paragraph{Output instability}
Given the same prompt, a model may generate inconsistent or opposing responses.   %
Such unstable behavior has high \riskharm\ and \riskfail, particularly in binary cases if opposing responses are both likely and in low \spec\ settings where a user will be less skeptical of potentially incorrect answers. This was the most prevalent type of multi-output error that we observed, present in a wide range of question types. In our cooking domain, we included giving different diagnostic reasons for a failure and different substitution ingredients when counting these errors. Figure \ref{ex:almondFlour} shows how some responses state that an almond flour substitute would work while others disagree, demonstrating output instability with high \riskfail\ due to opposing responses. In Figure \ref{ex:tahini} we see different quantities of sesame seeds required to make 1/2 cup of tahini paste, ranging from 2 tablespoons to 1.5 cups (24 tablespoons). Responding with a low quantity has high \riskfail, but \riskharm\ is low because the result of the failure is making not enough tahini paste, requiring the user to repeat the task with more sesame seeds. Responding with a higher quantity than is actually required has no \riskfail\ and low \riskharm\ because the result would be having excess tahini paste that can be stored for other uses.

    \paragraph{Recommendations}
    
     Procedures often call for specific brands of materials, and using different brands can have a large impact on the success of the procedure. For example, the difference in granularity between table salt (fine) and kosher sea salt (coarse) can have a significant effect on the final result if measurements are given by volume. \riskfail increases with different types and  coarseness of salt because the resulting dish could be too salty to consume. This error can also increase \riskharm, for example, when a brand recommendation is associated with durability, (e.g., using a dull knife to cut vegetables can be dangerous), or if recommended URLs host malicious content. Figure \ref{ex:tahini} illustrates this error with responses suggesting different recipes for making tahini paste. The creaminess of the paste and flavor, if one recipe uses roasted sesame seeds, can differ between recipes.

    \paragraph{Leading question agreement}
    
    Leading questions in ProcDocQA questions will contain suggested answers in the question, changing the intent of the question to both verification of existing knowledge and a request for new knowledge. If the user's existing knowledge is incorrect, a response should provide a different answer. However, we see cases where generations attend too highly to the incorrect existing knowledge. Figure \ref{ex:tahini} shows an example of this with the ``2-4 oil'' span in the original question. Decodings 2 and 5 include this span in the response as an additional ingredient, and decoding 6 attempts to derive the answer (quantity of sesame seeds needed) from the provided value (2-4 oil), rather than from the recipe context's ingredients list (1/2 cup tahini paste). \riskfail\ increases if the user's existing knowledge is incorrect and reinforced by the response.

    \paragraph{Hallucination}
    
    \begin{figure}
        \centering
        \includegraphics[width=.45\textwidth]{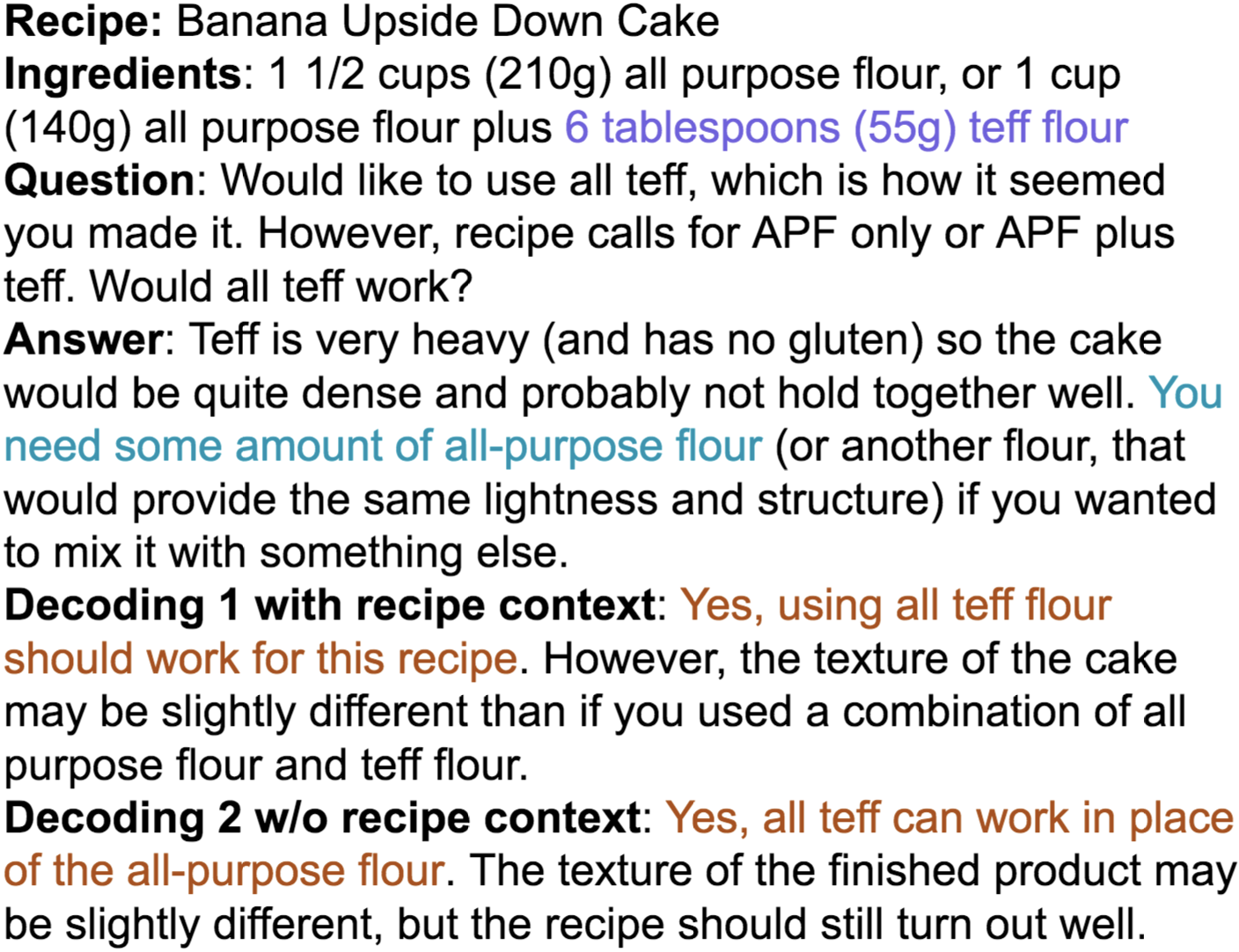}
        \caption{\textbf{Hallucination, language style behaviors.} The \textcolor{c_blue}{reference answer} states that you cannot use pure teff flour in this recipe, yet both decodings with and without any recipe context in the prompt state the \textcolor{c_orange}{opposite}.
        However, the recipe only \textit{optionally} uses teff at $\approx 25\%$ of the total flour content by weight, suggesting that you can't use all teff, regardless of any knowledge about the properties of teff (a dense gluten-free grain). }
        \label{ex:teff}
    \end{figure} 

    Many questions in ProcDocQA can seemingly be answered with a simple table lookup (e.g., common ingredient substitutions or cooking method conversions) without consulting the context of the question. 
    This is not often the case, as can be seen by the extensive work in automatically recommending ingredient substitutions (e.g., \citealp{Liu2018AlternativeIR,Ooi2015IngredientSR,Pacfico2021IngredientSR}). Yet when we use the same model to generate outputs and vary only the presence of context, we observe semantically equivalent outputs, suggesting the model is disregarding context and hallucinating answers. Tasks requiring higher \spec\ will have high \riskharm\ from hallucinations because higher \spec\ tasks require more environment-specific information.

    Figure \ref{ex:teff} shows how the decoding output is semantically similar regardless of whether recipe context is included. Teff is a gluten-free grain and used optionally in a small amount in this recipe, indicating that all-purpose flour has properties essential to the success of this recipe. Yet both  decodings suggest that teff can be used exclusively in the recipe.\footnote{Google's search engine returns results saying (incorrectly) that one can substitute all-purpose flour with teff, so it is understandable that the system propagates it.}
    
    \paragraph{Language style}
    Language style can influence \riskfail\ if there is a mismatch with the intended audience, such as when a context requires precise answers, the users have low \spec, and the language expressed is imprecise. \riskfail\ may increase due to the user believing there is room for error and not having the \spec\ to suspect higher precision is required. Domains with high \riskharm\ or high \spec\ requirements may generally prefer to have highly precise language that inspires confidence because there are higher requirements for the success of the procedure. For example, tempering chocolate is a cooking procedure calling for relatively high \spec, requiring the temperature of melted dark chocolate to be maintained between 86--90$^\circ$F. In contrast, slow cooking stew requires relatively low \spec, and instructions might suggest cooking for at minimum 5 hours, but anywhere up to 8 hours, with no impact on \riskfail. Figure \ref{ex:teff} decodings hedge, using less certain language when stating that using teff flour ``should'' or ``can'' work, indicating to the user that the substitution has some \riskfail. If model confidence in its answers is low relative to the \spec\ required for the question, then the model should refrain from responding (``Unable to provide a response''), in order to reduce \riskfail.

    \section{RADQ Post-Study Update} \label{sec:radq-redux} \label{sec:rad-poststudy} 
    
    \begin{figure}
    \centering
    \includegraphics[width=.45\textwidth]{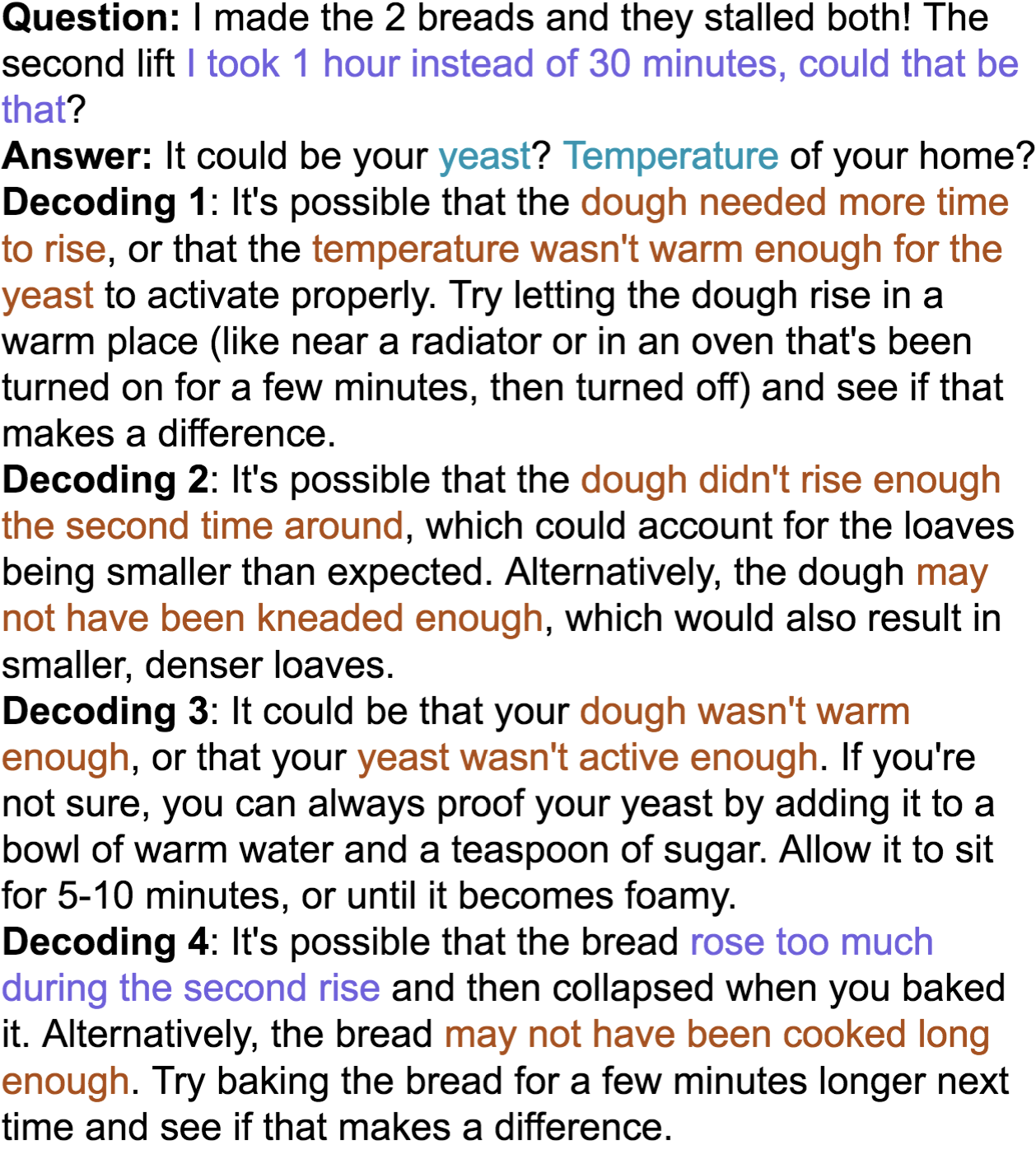}
    \caption{The \textcolor{c_blue}{reference answer} suggests problems with yeast health and rising environment temperature, whereas the decodings also suggest issues with \textcolor{c_orange}{rising time, gluten development in the kneading process, dough temperature, and cooking time}. }
    \label{ex:breadstall}
\end{figure}
    
    Informed by our user perspective study and multi-output error analysis, we update our RADQ responses from \S\ref{sec:radq-init} and connect to existing research that could help inform more risk-aware designs.
    
    \paragraph{Q2} Though we were initially skeptical when we observed explicit recommendations of specific brands in answers, users preferred them even with the knowledge that the recommendation comes from an automated system (Appendix \ref{app:survey-results}). Filtering recommendations might still be preferable if the system lacks knowledge of availability in the user's geographic locale or cannot verify the integrity of the recommendations because the user experience suffers and \riskfail increases if recommendations are inaccessible.  Work in QA answer verification (e.g., \citealp{wang-etal-2018-multi-passage}) and fact verification (e.g., \citealp{park-etal-2022-faviq}) where answers require citations could help filter such recommendations. Further work in balancing user preferences with theoretical harms of AI bias is needed to support development of practical, safe systems while maximizing user experience.
    
    \paragraph{Q3} We expected hedged responses to provide low satisfaction, yet this was not the case for either crowdworkers or experts: 79\% of answers with highest satisfaction contained hedging,\footnote{Hedging wordlist: \url{https://github.com/words/hedges}} 2\% higher than all other answers. We hypothesize there may be a perceptual gap in user understanding of the relationship between uncertainty and \riskfail, as well as domain norms at play---users are aware that cooking is not an exact science. Work in risk communication and management (e.g., \citealp{Renn1996,https://doi.org/10.1111/risa.12071}), where qualitative risk perception studies balance out quantitative risk models to guide risk communications, can help inform when using hedging is appropriate.
    
    \paragraph{Q2, Q5} GPT-3's frequent output instability within just 10 generations was surprisingly common, despite the case study (which used only the first generation) indicating the high quality/correctness of model output. This observation resurfaces questions in answer merging (\citealp{Gondek2012AFF,merging-ranking-answers-lopez-etal-2009}) with application-specific design decisions. Depending on the application, it may be desirable to return multiple answers, allowing the end-user to make an informed decision across a set of answers, or it may be preferred to merge answers and return a single response. Work in uncertainty visualization (e.g., \citealp{uncertainty-visualization}) can help inform how to present merged answers by drawing on the psychology of how different approaches are viewed. Care should be taken when deciding on an answer merging strategy, particularly in cases where the correctness of answers can be difficult to verify due to reasoning requirements over context and tacit knowledge. Figure \ref{ex:breadstall} describes many possible reasons for recipe failure, yet it is difficult to determine which, if any, of the possible reasons are correct for the specific user. Poorly chosen answer merging strategies and visual presentation of multiple results can confuse the user and increase \riskfail\ even if all presented answers are correct. 

    The second version of our recipe QA system may include:
    \begin{itemize}
        \item \spec estimator for recipes (which may already be provided), for calibrating language style edits (e.g., \citealp{august-etal-2022-generating,leroy2010influence})
        \item Question classifier to inform answer merging strategies and visualizations (e.g., \citealp{cortes-etal-2020-empirical}) 
        \item Answer merging strategies dependent on question types (e.g., \citealp{glockner2007logical})
        \item Multiple answer visualizations with uncertainty information and source verification for as many answers as possible (e.g., \citealp{ruckle-gurevych-2017-end})
        \item Recommendation filter to verify brand and URL integrity conditioned on availability of geographic information (e.g., \citealp{provos2008all})
    \end{itemize}

    \section{Conclusions}

    On the surface, vanilla GPT-3 presents itself as a powerful system ready for deployment as (among other things) a cooking recipe question answering system with no additional filtering or adaptation needed of its outputs. 
    However, multiple generations over the same question revealed several types of error with varying degrees of \riskharm and  \riskfail relative to \spec of the users. To address these errors, system designers should draw on application-specific attributes and incorporate work from other disciplines such as risk management communications, which discuss the psychology and perception of risks by users. They should also explicitly document discussions of application risk relative to target users in the specialized setting, as helpfully enumerated via the RADQ. 
    Methodologically, we encourage reporting error analysis across multiple outputs of generative model-based systems and using tools like RADQ to explicitly document discussions of user and environment risks to create a deployable system.

    \section{Limitations}
    Cooking recipes constitute a single genre within ProcDocQA, with a well-grounded task and large range in \riskharm\ and user \spec. Our case study only investigated a narrow range in \riskharm\ and \spec\ due to the nature of the data: self-published blog recipes in English collected with simple heuristics.

    The first version of RADQ was informed by theoretical AI risk frameworks and our CookingQA case study; we anticipate the questionnaire evolving greatly when informed by other QA domains with different levels of \riskharm\ and \spec. This work only considers immediate risks to humans; longitudinal risks such as the propagation of information are an open research topic.

    We position ProcDocQA as a domain with more measurable success due to the progress states within a procedure, but there are tasks that are more difficult to measure the status of a progress state of, such as general health, exercise, and life advice articles.

    This work contributes to risk mitigation by concretizing risks in user-aware scenarios. Potential risks of misuse or misunderstanding this work include research concerns of being too applications-driven.

    \section{Ethics Statement}
    
    User studies were conducted after review by our institution's IRB, and participants were paid a fair wage in accordance with our local government. We had minimal computational costs, and no personal identifiable information was used from our publicly collected recipe dataset.

\bibliographystyle{acl_natbib}

\begin{thebibliography}{35}
\expandafter\ifx\csname natexlab\endcsname\relax\def\natexlab#1{#1}\fi

\bibitem[{August et~al.(2022)August, Reinecke, and Smith}]{august-etal-2022-generating}
Tal August, Katharina Reinecke, and Noah~A. Smith. 2022.
\newblock \href {https://doi.org/10.18653/v1/2022.acl-long.569} {Generating scientific definitions with controllable complexity}.
\newblock In \emph{Proceedings of the 60th Annual Meeting of the Association for Computational Linguistics (Volume 1: Long Papers)}, pages 8298--8317, Dublin, Ireland. Association for Computational Linguistics.

\bibitem[{Bender and Friedman(2018)}]{bender-friedman-2018-data}
Emily~M. Bender and Batya Friedman. 2018.
\newblock \href {https://doi.org/10.1162/tacl_a_00041} {Data statements for natural language processing: Toward mitigating system bias and enabling better science}.
\newblock \emph{Transactions of the Association for Computational Linguistics}, 6:587--604.

\bibitem[{Bier and Lin(2013)}]{https://doi.org/10.1111/risa.12071}
Vicki~M. Bier and Shi-Woei Lin. 2013.
\newblock \href {https://doi.org/https://doi.org/10.1111/risa.12071} {On the treatment of uncertainty and variability in making decisions about risk}.
\newblock \emph{Risk Analysis}, 33(10):1899--1907.

\bibitem[{Blodgett et~al.(2020)Blodgett, Barocas, Daum{\'e}~III, and Wallach}]{blodgett-etal-2020-language}
Su~Lin Blodgett, Solon Barocas, Hal Daum{\'e}~III, and Hanna Wallach. 2020.
\newblock \href {https://doi.org/10.18653/v1/2020.acl-main.485} {Language (technology) is power: A critical survey of {``}bias{''} in {NLP}}.
\newblock In \emph{Proceedings of the 58th Annual Meeting of the Association for Computational Linguistics}, pages 5454--5476, Online. Association for Computational Linguistics.

\bibitem[{Brown et~al.(2020)Brown, Mann, Ryder, Subbiah, Kaplan, Dhariwal, Neelakantan, Shyam, Sastry, Askell, Agarwal, Herbert-Voss, Krueger, Henighan, Child, Ramesh, Ziegler, Wu, Winter, Hesse, Chen, Sigler, Litwin, Gray, Chess, Clark, Berner, McCandlish, Radford, Sutskever, and Amodei}]{gpt3}
Tom~B. Brown, Benjamin Mann, Nick Ryder, Melanie Subbiah, Jared Kaplan, Prafulla Dhariwal, Arvind Neelakantan, Pranav Shyam, Girish Sastry, Amanda Askell, Sandhini Agarwal, Ariel Herbert-Voss, Gretchen Krueger, Tom Henighan, Rewon Child, Aditya Ramesh, Daniel~M. Ziegler, Jeffrey Wu, Clemens Winter, Christopher Hesse, Mark Chen, Eric Sigler, Mateusz Litwin, Scott Gray, Benjamin Chess, Jack Clark, Christopher Berner, Sam McCandlish, Alec Radford, Ilya Sutskever, and Dario Amodei. 2020.
\newblock \href {http://arxiv.org/abs/2005.14165} {Language models are few-shot learners}.

\bibitem[{Buiten(2019)}]{buiten_2019}
Miriam~C Buiten. 2019.
\newblock \href {https://doi.org/10.1017/err.2019.8} {Towards intelligent regulation of artificial intelligence}.
\newblock \emph{European Journal of Risk Regulation}, 10(1):41–59.

\bibitem[{Clark et~al.(2021)Clark, August, Serrano, Haduong, Gururangan, and Smith}]{Clark2021Allt}
Elizabeth Clark, Tal August, Sofia Serrano, Nikita Haduong, Suchin Gururangan, and Noah~A. Smith. 2021.
\newblock All that’s ‘human’ is not gold: Evaluating human evaluation of generated text.
\newblock In \emph{Annual Meeting of the Association for Computational Linguistics}.

\bibitem[{Cortes et~al.(2020)Cortes, Woloszyn, Binder, Himmelsbach, Barone, and M{\"o}ller}]{cortes-etal-2020-empirical}
Eduardo Cortes, Vinicius Woloszyn, Arne Binder, Tilo Himmelsbach, Dante Barone, and Sebastian M{\"o}ller. 2020.
\newblock \href {https://aclanthology.org/2020.lrec-1.665} {An empirical comparison of question classification methods for question answering systems}.
\newblock In \emph{Proceedings of the Twelfth Language Resources and Evaluation Conference}, pages 5408--5416, Marseille, France. European Language Resources Association.

\bibitem[{Dhuliawala et~al.(2022)Dhuliawala, Adolphs, Das, and Sachan}]{Dhuliawala2022CalibrationOM}
Shehzaad Dhuliawala, Leonard Adolphs, Rajarshi Das, and Mrinmaya Sachan. 2022.
\newblock Calibration of machine reading systems at scale.
\newblock \emph{ArXiv}, abs/2203.10623.

\bibitem[{Dou et~al.(2021)Dou, Forbes, Koncel-Kedziorski, Smith, and Choi}]{dou2021scarecrow}
Yao Dou, Maxwell Forbes, Rik Koncel-Kedziorski, Noah~A. Smith, and Yejin Choi. 2021.
\newblock \href {http://arxiv.org/abs/2107.01294} {Scarecrow: A framework for scrutinizing machine text}.

\bibitem[{Ganguli et~al.(2022)Ganguli, Hernandez, Lovitt, Askell, Bai, Chen, Conerly, Dassarma, Drain, Elhage, El~Showk, Fort, Hatfield-Dodds, Henighan, Johnston, Jones, Joseph, Kernian, Kravec, Mann, Nanda, Ndousse, Olsson, Amodei, Brown, Kaplan, McCandlish, Olah, Amodei, and Clark}]{predictability_surprise_lm}
Deep Ganguli, Danny Hernandez, Liane Lovitt, Amanda Askell, Yuntao Bai, Anna Chen, Tom Conerly, Nova Dassarma, Dawn Drain, Nelson Elhage, Sheer El~Showk, Stanislav Fort, Zac Hatfield-Dodds, Tom Henighan, Scott Johnston, Andy Jones, Nicholas Joseph, Jackson Kernian, Shauna Kravec, Ben Mann, Neel Nanda, Kamal Ndousse, Catherine Olsson, Daniela Amodei, Tom Brown, Jared Kaplan, Sam McCandlish, Christopher Olah, Dario Amodei, and Jack Clark. 2022.
\newblock \href {https://doi.org/10.1145/3531146.3533229} {Predictability and surprise in large generative models}.
\newblock In \emph{2022 ACM Conference on Fairness, Accountability, and Transparency}, FAccT '22, page 1747–1764, New York, NY, USA. Association for Computing Machinery.

\bibitem[{Gl{\"o}ckner et~al.(2007)Gl{\"o}ckner, Hartrumpf, and Leveling}]{glockner2007logical}
Ingo Gl{\"o}ckner, Sven Hartrumpf, and Johannes Leveling. 2007.
\newblock Logical validation, answer merging and witness selection-a study in multi-stream question answering.
\newblock In \emph{RIAO}, pages 758--777.

\bibitem[{Gondek et~al.(2012)Gondek, Lally, Kalyanpur, Murdock, Duboue, Zhang, Pan, Qiu, and Welty}]{Gondek2012AFF}
David Gondek, Adam Lally, Aditya Kalyanpur, J.~William Murdock, Pablo Duboue, Lei Zhang, Yue Pan, Zhaoming Qiu, and Chris Welty. 2012.
\newblock A framework for merging and ranking of answers in deepqa.
\newblock \emph{IBM J. Res. Dev.}, 56:14.

\bibitem[{Grandstrand(2022)}]{uncertainty-visualization}
Ove Grandstrand. 2022.
\newblock Uncertainty visualization.
\newblock In JWalter~W. Piegorsch, Richard~A. Levine, Hao~Helen Zhang, and Thomas C.~M. Lee, editors, \emph{Computational Statistics in Data Science}, chapter~22, pages 405--421. Wiley, Oxford.

\bibitem[{Jacobs and Wallach(2021)}]{Jacobs2021MeasurementAF}
Abigail~Z. Jacobs and Hanna~M. Wallach. 2021.
\newblock Measurement and fairness.
\newblock \emph{Proceedings of the 2021 ACM Conference on Fairness, Accountability, and Transparency}.

\bibitem[{Lee et~al.(2020)Lee, Floridi, and Denev}]{Lee2020InnovatingWC}
Michelle Seng~Ah Lee, L.~Floridi, and Alexander Denev. 2020.
\newblock Innovating with confidence - embedding ai governance and fairness in a financial services risk management framework.
\newblock \emph{Social Science Research Network}.

\bibitem[{Leroy et~al.(2010)Leroy, Helmreich, and Cowie}]{leroy2010influence}
Gondy Leroy, Stephen Helmreich, and James~R Cowie. 2010.
\newblock The influence of text characteristics on perceived and actual difficulty of health information.
\newblock \emph{International journal of medical informatics}, 79(6):438--449.

\bibitem[{Liu et~al.(2018)Liu, Chen, Lai, Wu, and Wei}]{Liu2018AlternativeIR}
Kuan-Hung Liu, Hung-Chih Chen, Kuan-Ting Lai, Yi-Ying Wu, and Chih-Ping Wei. 2018.
\newblock Alternative ingredient recommendation: A co-occurrence and ingredient category importance based approach.
\newblock In \emph{PACIS}.

\bibitem[{Lopez et~al.(2009)Lopez, Nikolov, Fernandez, Sabou, Uren, and Motta}]{merging-ranking-answers-lopez-etal-2009}
Vanessa Lopez, Andriy Nikolov, Miriam Fernandez, Marta Sabou, Victoria Uren, and Enrico Motta. 2009.
\newblock Merging and ranking answers in the semantic web: The wisdom of crowds.
\newblock In \emph{The Semantic Web}, pages 135--152, Berlin, Heidelberg. Springer Berlin Heidelberg.

\bibitem[{Manheim and Kaplan(2019)}]{privacy_democracy}
Karl Manheim and Lyric Kaplan. 2019.
\newblock \href {https://heinonline.org/HOL/Page?handle=hein.journals/yjolt21&div=4&id=&page=&collection=journals} {Artificial intelligence: Risks to privacy and democracy}.
\newblock \emph{Yale Journal of Law and Technology}, 21:106--188.

\bibitem[{Ooi et~al.(2015)Ooi, Iiba, and Takano}]{Ooi2015IngredientSR}
Ami Ooi, Toshiya Iiba, and Kosuke Takano. 2015.
\newblock Ingredient substitute recommendation for allergy-safe cooking based on food context.
\newblock \emph{2015 IEEE Pacific Rim Conference on Communications, Computers and Signal Processing (PACRIM)}, pages 444--449.

\bibitem[{Pac{\'i}fico et~al.(2021)Pac{\'i}fico, Britto, and Ludermir}]{Pacfico2021IngredientSR}
Luciano Demetrio~Santos Pac{\'i}fico, Larissa F.~S. Britto, and Teresa~B Ludermir. 2021.
\newblock Ingredient substitute recommendation based on collaborative filtering and recipe context for automatic allergy-safe recipe generation.
\newblock \emph{Proceedings of the Brazilian Symposium on Multimedia and the Web}.

\bibitem[{Park et~al.(2022)Park, Min, Kang, Zettlemoyer, and Hajishirzi}]{park-etal-2022-faviq}
Jungsoo Park, Sewon Min, Jaewoo Kang, Luke Zettlemoyer, and Hannaneh Hajishirzi. 2022.
\newblock \href {https://doi.org/10.18653/v1/2022.acl-long.354} {{F}a{VIQ}: {FA}ct verification from information-seeking questions}.
\newblock In \emph{Proceedings of the 60th Annual Meeting of the Association for Computational Linguistics (Volume 1: Long Papers)}, pages 5154--5166, Dublin, Ireland. Association for Computational Linguistics.

\bibitem[{Provos et~al.(2008)Provos, Mavrommatis, Rajab, and Monrose}]{provos2008all}
Niels Provos, Panayiotis Mavrommatis, Moheeb Rajab, and Fabian Monrose. 2008.
\newblock All your iframes point to us.

\bibitem[{Raso et~al.(2018)Raso, Hilligoss, Krishnamurthy, Bavitz, and Levin}]{berkman_humanrights}
Filippo Raso, Hannah Hilligoss, Vivek Krishnamurthy, Christopher Bavitz, and Kim Levin. 2018.
\newblock \href {https://doi.org/http://dx.doi.org/10.2139/ssrn.3259344} {Artificial intelligence \& human rights: Opportunities \& risks}.

\bibitem[{Renn et~al.(1996)Renn, Webler, and Kastenholz}]{Renn1996}
Ortwin Renn, Thomas Webler, and Hans Kastenholz. 1996.
\newblock \href {https://doi.org/10.1007/978-94-015-8619-1_7} {\emph{Perception of Uncertainty: Lessons for Risk Management and Communication}}, pages 163--181. Springer Netherlands, Dordrecht.

\bibitem[{R{\"u}ckl{\'e} and Gurevych(2017)}]{ruckle-gurevych-2017-end}
Andreas R{\"u}ckl{\'e} and Iryna Gurevych. 2017.
\newblock \href {https://aclanthology.org/P17-4004} {End-to-end non-factoid question answering with an interactive visualization of neural attention weights}.
\newblock In \emph{Proceedings of {ACL} 2017, System Demonstrations}, pages 19--24, Vancouver, Canada. Association for Computational Linguistics.

\bibitem[{Straw and Callison-Burch(2020)}]{mental_health_biases}
Isabel Straw and Chris Callison-Burch. 2020.
\newblock \href {https://doi.org/10.1371/journal.pone.0240376} {Artificial intelligence in mental health and the biases of language based models}.
\newblock \emph{PLOS ONE}, 15(12):1--19.

\bibitem[{Su et~al.(2019)Su, Guo, Fan, Lan, and Cheng}]{risk_webqa}
Lixin Su, Jiafeng Guo, Yixin Fan, Yanyan Lan, and Xueqi Cheng. 2019.
\newblock \href {https://doi.org/10.1145/3331184.3331261} {Controlling risk of web question answering}.
\newblock In \emph{Proceedings of the 42nd International ACM SIGIR Conference on Research and Development in Information Retrieval}, SIGIR'19, page 115–124, New York, NY, USA. Association for Computing Machinery.

\bibitem[{Tan et~al.(2021)Tan, Joty, Baxter, Taeihagh, Bennett, and Kan}]{Tan2021ReliabilityTF}
Samson Tan, Shafiq~R. Joty, K.~Baxter, Araz Taeihagh, G.~Bennett, and Min-Yen Kan. 2021.
\newblock Reliability testing for natural language processing systems.
\newblock In \emph{Annual Meeting of the Association for Computational Linguistics}.

\bibitem[{Tan et~al.(2022)Tan, Taeihagh, and Baxter}]{Tan2022TheRO}
Samson Tan, Araz Taeihagh, and K.~Baxter. 2022.
\newblock The risks of machine learning systems.
\newblock \emph{ArXiv}, abs/2204.09852.

\bibitem[{Wang et~al.(2018)Wang, Liu, Liu, He, Lyu, Wu, Li, and Wang}]{wang-etal-2018-multi-passage}
Yizhong Wang, Kai Liu, Jing Liu, Wei He, Yajuan Lyu, Hua Wu, Sujian Li, and Haifeng Wang. 2018.
\newblock \href {https://doi.org/10.18653/v1/P18-1178} {Multi-passage machine reading comprehension with cross-passage answer verification}.
\newblock In \emph{Proceedings of the 56th Annual Meeting of the Association for Computational Linguistics (Volume 1: Long Papers)}, pages 1918--1927, Melbourne, Australia. Association for Computational Linguistics.

\bibitem[{Wang and Ai(2021)}]{Wang2021ControllingTR}
Zhenduo Wang and Qingyao Ai. 2021.
\newblock Controlling the risk of conversational search via reinforcement learning.
\newblock \emph{Proceedings of the Web Conference 2021}.

\bibitem[{Wu et~al.(2021)Wu, Shen, Li, Zhou, and Lu}]{Wu2021ABF}
Zongda Wu, Shigen Shen, Huxiong Li, Haiping Zhou, and Chenglang Lu. 2021.
\newblock A basic framework for privacy protection in personalized information retrieval: An effective framework for user privacy protection.
\newblock \emph{J. Organ. End User Comput.}, 33:1--26.

\bibitem[{Zhang et~al.(2022)Zhang, Chan, Yan, and Bose}]{ZHANG2022113800}
Xiaoge Zhang, Felix~T.S. Chan, Chao Yan, and Indranil Bose. 2022.
\newblock \href {https://doi.org/https://doi.org/10.1016/j.dss.2022.113800} {Towards risk-aware artificial intelligence and machine learning systems: An overview}.
\newblock \emph{Decision Support Systems}, 159:113800.

\end{thebibliography}

\appendix

\label{sec:appendix}

\clearpage
\section{Data}
\subsection{Cooking Dataset}
\label{app:uqa--details}
The custom dataset collected for finetuning UnifiedQA consisted of 105k recipes from 192 blogs extracted from CommonCrawl accessed on July 29, 2022. Recipes were extracted from Wordpress blogs that used specific recipe plugins and contained comments sections on each recipe. Question-answer pairs were mined from the comments sections using simple heuristics: 1) does the comment contain common question n-grams (\textit{who, what, where, when, how, instead, substitute,         substitution, replace, replacement, changes, why, can i, can you}), and 2) the first reply to a question comment is the answer.

\begin{table*}[ht]
    \centering
    \begin{tabular}{p{0.95\linewidth} }
    \toprule
      1 1/4 lbs butternut squash diced 1-inch \newline
1 tbsp oil\newline
4 cloves garlic , smashed with the side of a knife\newline
1/4 cup ricotta , I prefer Polly-o\newline
1/4 cup Pecorino Romano , plus optional more for serving\newline
1/4 teaspoon kosher and black pepper , to taste\newline
1/4 teaspoon nutmeg\newline
24 square wonton wrappers\newline
1 large egg , beaten\newline
2 tablespoons salted butter\newline
8 fresh sage leaves , divided\newline
Preheat the oven to 400F. Place butternut, 4 sage leaves and garlic on a sheet pan and toss with 1 tablespoon oil. Season with 1/4 teaspoon salt and pepper, to taste. Roast until tender, about 35 minutes. Transfer to a bowl and mash with a fork until very smooth (a blender would work too). Mix in ricotta and pecorino, season with nutmeg, 1/4 teaspoon salt and black pepper. Place the wonton wrapper on a work surface, brush the edge lightly with egg wash and add 1 tablespoon filling onto the center. Fold over into a triangle and press the edges to seal. Cover with a damp cloth while you make the rest. Chop remaining sage leaves. Place butter and sage in a medium saucepan and melt over low heat. Keep warm over very low heat. Bring a large pot of salted water to a boil. Add half of the ravioli (they are very delicate) and cook until the rise to the surface, about 2 minutes. Use a slotted spoon to remove and add to the pan with the butter. Repeat with the remaining ravioli. Gently toss raviolis with the butter until warm, 1 to 2 minutes. Top with black pepper and serve with additional Pecorino Romano, if desired.\newline
\textbf{Question}: Hello!!! Can you use frozen butternut squash that’s already cubed? (Just to save time so I don’t have to peel, cube myself?) if so, how would you recommend going about it?\newline
\textbf{Answer}: \\ \hline
    1 tablespoon olive oil\newline
    2 skinless and boneless chicken breast fillets , halved horizontally to make four fillets\newline
    Salt , to season\newline
    14 ounces | 400 grams sliced mushrooms\newline
    2 teaspoons butter\newline
    1 large french shallot , finely chopped (normal shallot for U.S readers)\newline
    1/2 cup (about 130ml) champagne (or sparkling white wine)\newline
    2/3 cup milk (or heavy / thickened cream)\newline
    1 teaspoon of cornstarch (corn flour) -- only if using milk\newline
    Fresh chopped parsley , to garnish
Heat the olive oil in a skillet of pan over medium heat. Season each chicken filet with a pinch of salt. Sear chicken on both sides, for about 3-5 minutes each side (depending on thickness), until golden all over. Transfer chicken to a plate. Fry the mushrooms in the butter and fry for a further 3-5 minutes, or until just beginning to soften. Transfer to the same plate as the chicken. Cover and keep warm. Add the shallot into the pan and cook for 4 minutes, while occasionally stirring. Pour in the champagne; stir well, while scraping any food bits from the bottom of the pan for added flavour. Season with salt and pepper and allow to simmer for about 10 minutes. Add milk (or cream) and simmer for 2-3 minutes. If using milk: mix the cornstarch with a tablespoon of water; pour into the centre of the pan and mix throughout the sauce until slightly thickened. Add the chicken and mushrooms into the sauce along with any cooking juices. Coat in the creamy sauce. Sprinkle with parsley and serve!\newline
\textbf{Question}: Do you have any suggestions for a mushroom substitute? That’s a recipe deal breaker for me, otherwise this looks great.\newline
\textbf{Answer}: \\
\bottomrule
    \end{tabular}
    \caption{Example GPT-3 prompt(s). }
    \label{tab:gpt3PromptsContext}
\end{table*}

\section{Annotation Task} \label{app:caseStudy}
\label{app:categoriesImprovement}

The categories annotators could select for how a response could be improved:
\paragraph{Concise}  Ingredient quantities, 
Cooking temperatures/times, 
Nutritional information, 
Technical details, 
Expected intermediate or final results, 
States the recipe has been updated without specifying how in the answer 

\paragraph{Verbose} Ingredient quantities, 
Cooking temperatures/times, 
Nutritional information, 
Technical details, 
Expected intermediate or final results, 
Made up details/processes (that may be irrelevant) 

\paragraph{Misc.} Is too dependent on the specific setting (e.g., ingredients, tools, and skills) of the person following the recipe, 
Does not inspire confidence in whether to trust the answer to the question

Every category had an ``Other'' options where annotators could add custom categories.

\begin{table*}[htb]
    \centering
    \resizebox{\linewidth}{!}{
    \small
    \begin{tabular}{R{0.12\linewidth}L{0.30\linewidth}L{0.25\linewidth}L{0.25\linewidth}}
    \toprule
    Recipe Title & Question & Answer - Human & Answer - GPT3 \\ 
    \midrule
    \textit{Healthy Crock Pot Chicken Vegetable Soup} & Hi! I just made this over night last night...it smells delicious and I am very excited to eat it...however there is a lot floating around in there...looks curdled, but I know that it is not as there is no dairy in there. I still want to eat it...is it OK to eat, and is this supposed to happen? Thank you! & That is coagulated protein from the chicken breasts cooking. It seems to happen more with older or previously frozen chicken. It looks ugly but it is harmless and has almost not effect on taste. Skim it or ignore it. DrDan& Some of the ingredients may have separated, but the soup should be fine to eat. \\ \addlinespace[0.3cm]
    \textit{Grandmas Coleslaw} & How many bags of pre-cut cabbage would be needed for this recipe? Thanks, looks great! & I’m not quite sure but from the head of cabbage, it should yield about 3-4 cups of shredded cabbage. & You would need 4 bags of pre-cut cabbage for this recipe. \\ \addlinespace[0.3cm]
    \textit{Pineapple Upside Down Sugar Cookie Cakes} & How long can you store the cookies or freeze them? & I haven’t frozen them. As for storing, let the cookies cool completely and then place them in an airtight container. If doing a double layer, put a sheet of parchment paper or wax paper between the layers to prevent sticking. Seal the lid, and they can sit on the counter- best eaten within 2-3 days. & The cookies can be stored in an airtight container at room temperature for up to 1 week.  \\ \addlinespace[0.3cm]
    \textit{Triple Berry Cake} & I bought this from SLJ for my daughters biryhday ans everybody loved it! I will try to make this with your recipe. Do you know how many calories in the whole cake and what is its weight? Many thanks!& Hi Ray, unfortunately, I don’t have that information. Hope you give this a try anyway! Thank you so much for stopping by.& The whole cake has approximately 3,600 calories and weighs approximately 12 pounds.       \\ \addlinespace[0.3cm]
    \textit{Crock Pot Maple Dijon Chicken Drumsticks} & Does anyone know how I could convert this recipe to work in an electric pressure cooker (Instant Pot)? \hfil & I would just brown the chicken and set the manual pressure to 15 mins or use the poultry function if you have the 7 in 1 instant pot. & To convert this recipe to work in an electric pressure cooker, you will need to reduce the amount of liquid in the recipe. Maple syrup, balsamic vinegar, and Dijon mustard are all quite thick, so you may need to reduce the amount of each by half. You will also need to reduce the cooking time. Chicken legs usually cook through in about 30 minutes in an electric pressure cooker, so you will need to adjust the cooking time accordingly. \\ \addlinespace[0.3cm]
    \textit{Keto Tomato Tart – Low Carb} & First, Happy anniversary, glad you were able to escape and recharge a bit! Second, I’d love to try this base as a dessert; do you think I could do some subs and top it with thinly sliced apple? any specific suggestions? Last, almond flour is sooo expensive, I don’t have a Vitamix, I guess I could try my food processor or bkender Have you found a source where it’s less than \$12/lb? & 
    Thanks Karen! Yes, you can definitely use this crust for a dessert, I used it for my blueberry crostada in the Fall Recipes Ezine and it works perfectly! Any fruit that you’d use in a pie would work well, including thinly sliced apples. I’m working on a slab pie recipe this week so stay tuned for that. Almond flour IS expensive but for the best results in baking low carb desserts you really have to buy the blanched and finely ground good stuff. I get it on Amazon and buy it in 3 or 5 lb bags to get the best price – around \$10 per pound. Totally worth it! http://amzn.to/1KRiebQ & Yes, I think you could definitely use this base as a dessert crust! I would suggest using a sweeter cheese like mascarpone or cream cheese in the filling, and topping with thinly sliced apples and a drizzle of honey. As for the almond flour, I have found that Trader Joe's carries it for a pretty reasonable price. Hope this helps!\\
    \bottomrule
   
    \end{tabular}
    }
    \caption{Sample questions and answers from the recipes dataset.}
\end{table*}

\subsection{Annotator Recruitment}
Eight crowdworkers were recruited on Amazon Mechanical Turk (AMT). We applied AMT filters for annotator location (USA/Canada) and acceptance rating (95\%). Three experts were recruited from culinary training programs. Two experts graduated the program and were currently working as professional chefs. One expert was still in the program. Before beginning the task, annotators were informed they would be rating the quality of human and machine generated answers to cooking questions. They gave consent to participate in the experiment and could withdraw at any time.
\clearpage

\label{app:presurvey}

\begin{figure}[hb]
    \centering
    \includegraphics[width=.45\textwidth]{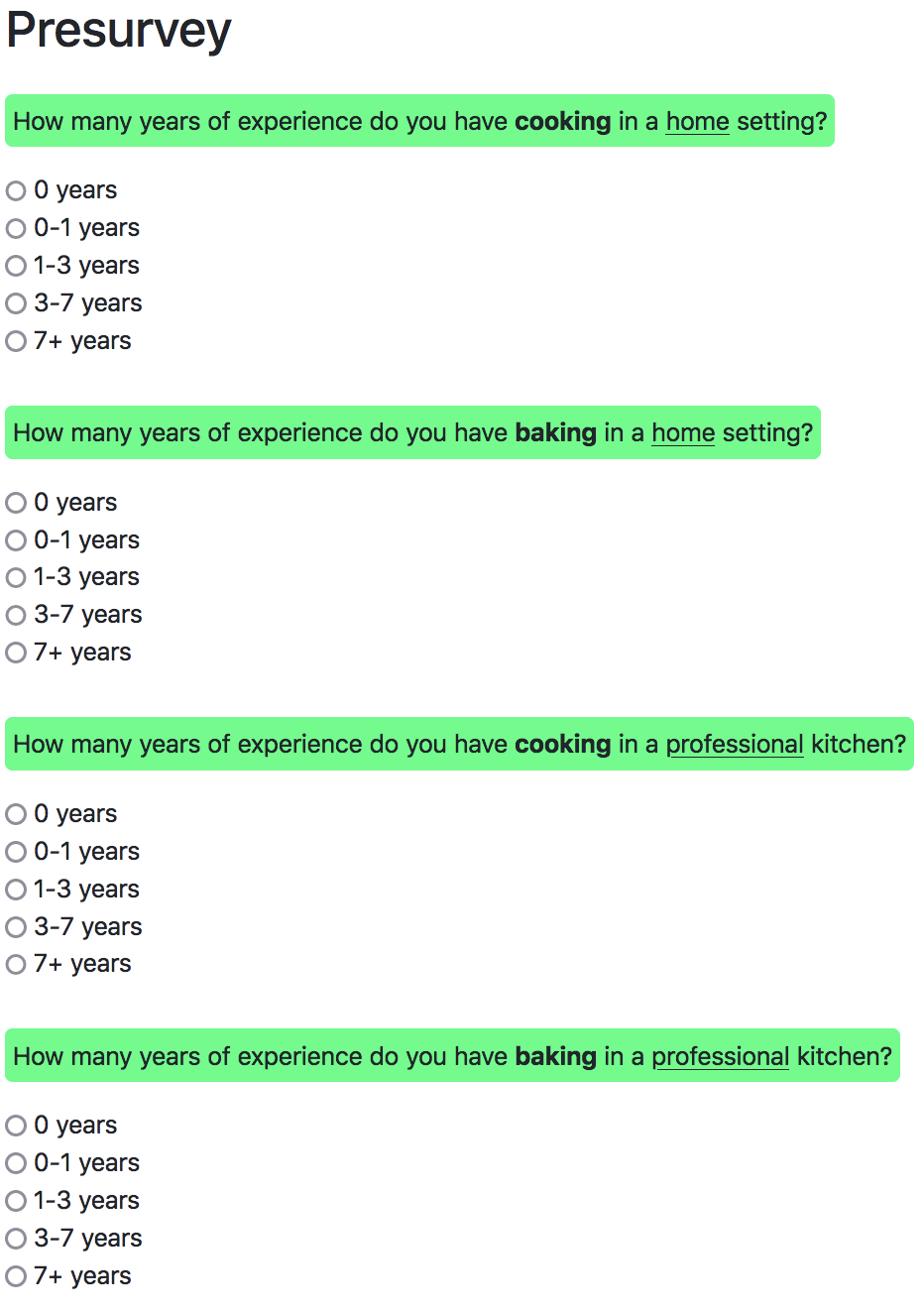}
    \label{fig:pre1}
\end{figure}

\begin{figure}[!h]
    \includegraphics[width=.45\textwidth,]{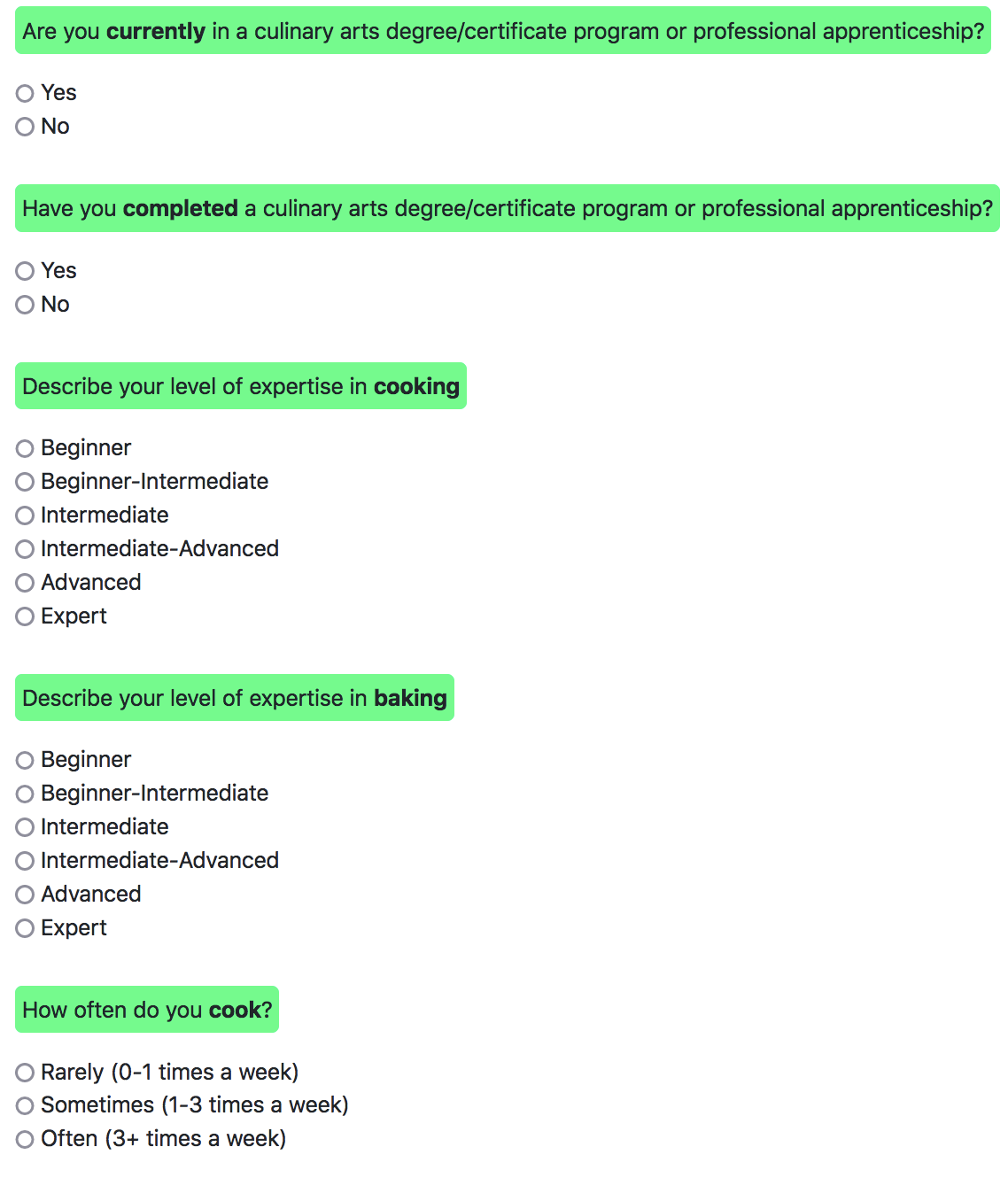}
    \vspace*{-5mm}
    \label{fig:pre2}
    \end{figure}
\begin{figure}[ht!]
    \includegraphics[width=.45\textwidth]{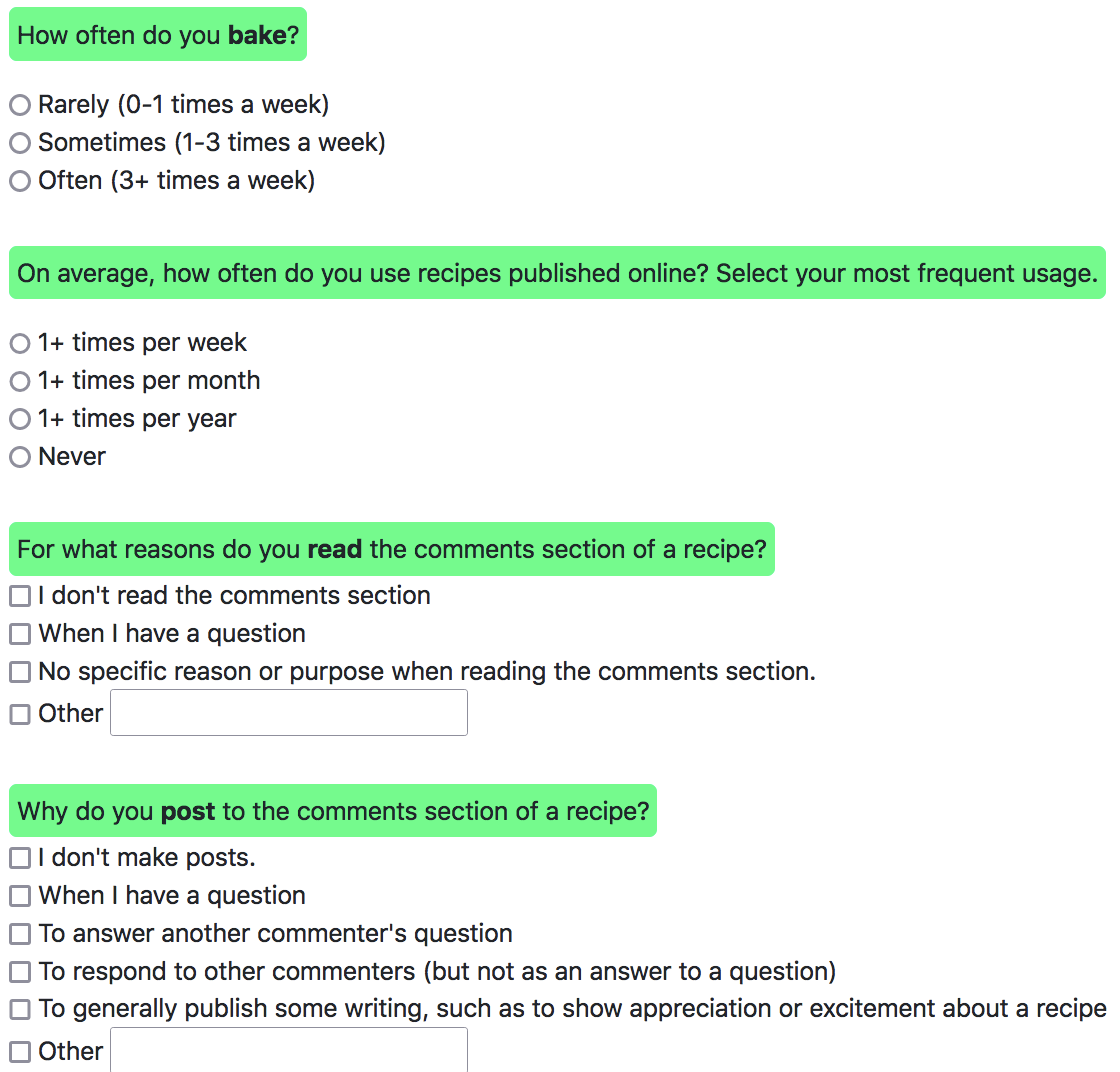}
    \label{fig:pre3}
    \end{figure}
\begin{figure}[ht!]
    \includegraphics[width=.45\textwidth]{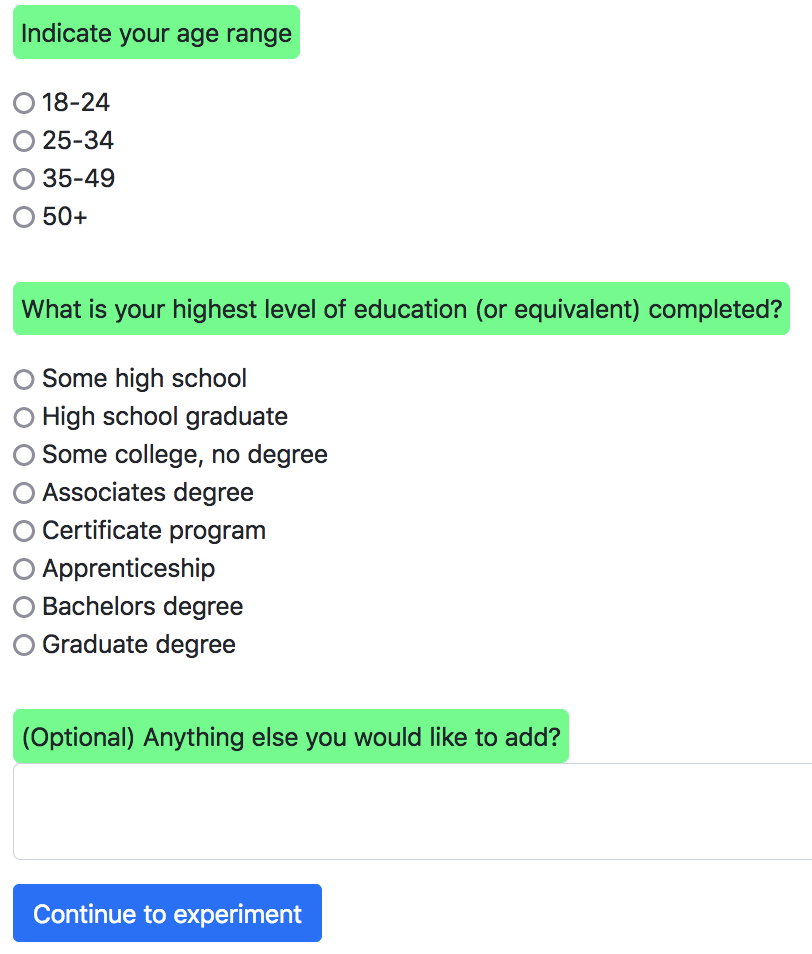}
    \caption{Presurvey questions.}
    \label{fig:pre4}
\end{figure}

\label{app:annotation_scheme}
\label{fig:appAnnotationTask}
\begin{figure}[!ht]
    \centering
    \vspace{-15mm}
    \includegraphics[width=.45\textwidth, scale=0.01]{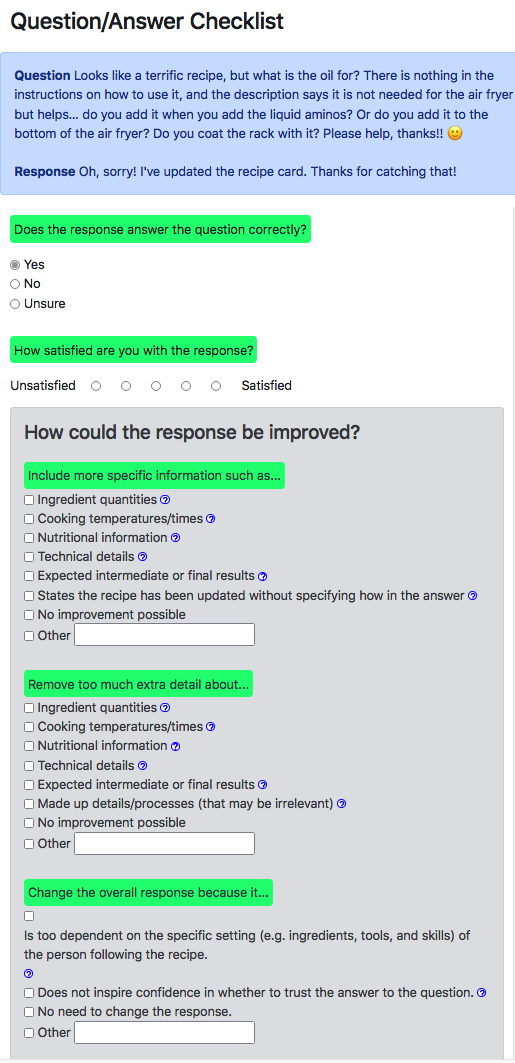}
    \caption{Screen shown when a response answers the question.}
    \label{fig:yes1}
\end{figure}
\begin{figure}[!bh]
    \centering
    \includegraphics[width=.45\textwidth, scale=0.01]{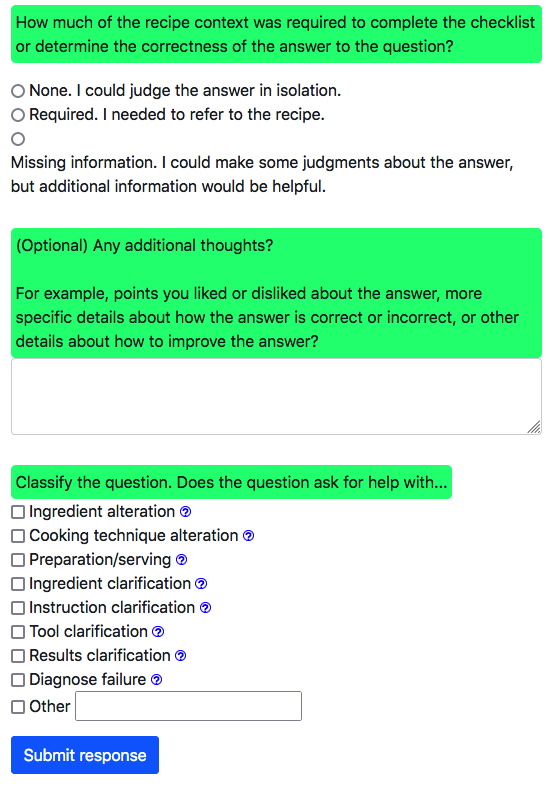}
    \caption{Screen shown when annotators state a response does and does not answer the question.}
    \label{fig:yesno}
\end{figure}
\clearpage
\begin{figure}[!htb]
    \centering
    \includegraphics[width=.45\textwidth]{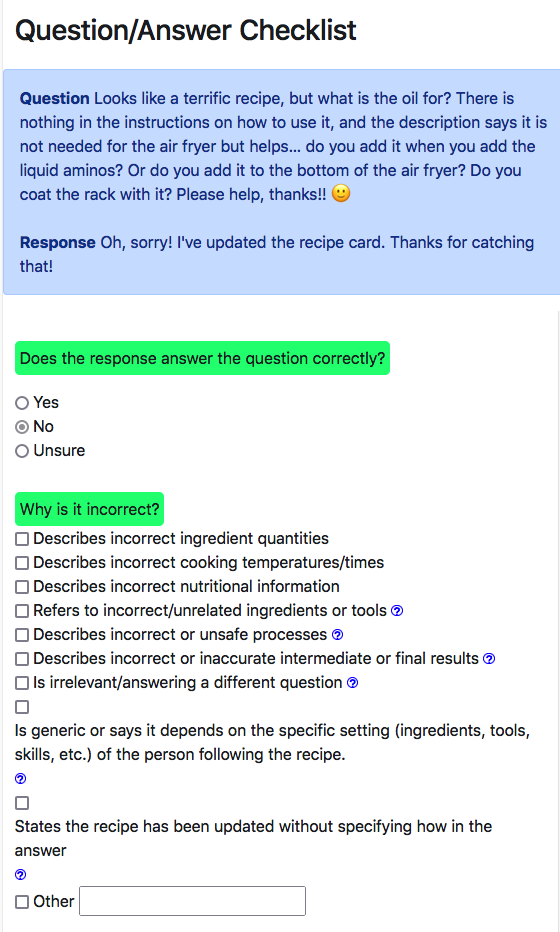}
    \caption{Screen shown when the response does not answer the questions.}
    \label{fig:no1}
\end{figure}
\begin{figure}[htb]
    \centering
    \includegraphics[width=.45\textwidth]{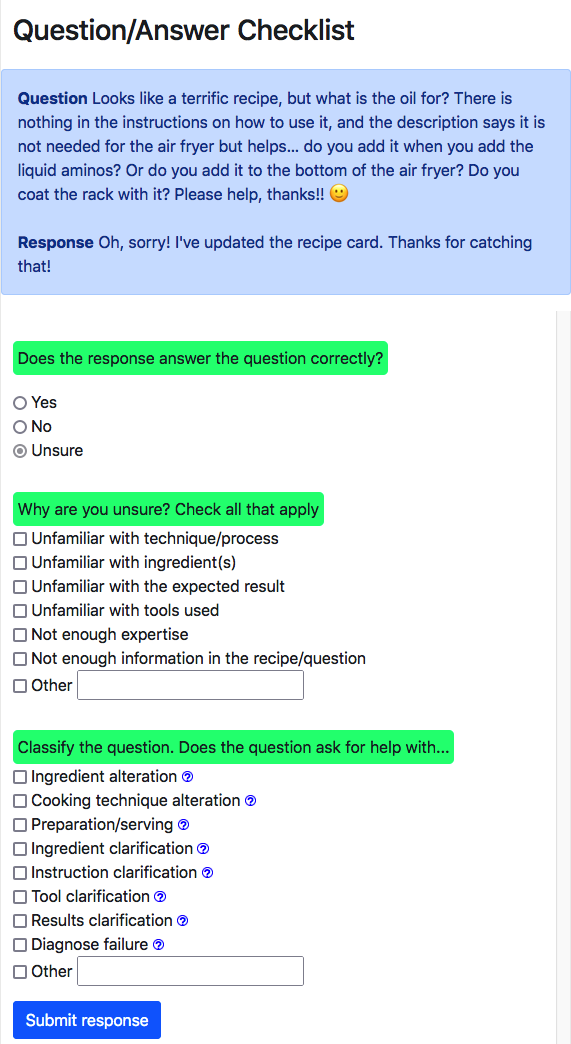}
    \caption{Screen shown if the annotator is unsure if the response answers the question.}
    \label{fig:unsure}
\end{figure}
\clearpage
\label{sec:postsurvey}
\begin{figure}[!ht]
    \centering
    \includegraphics[width=.45\textwidth]{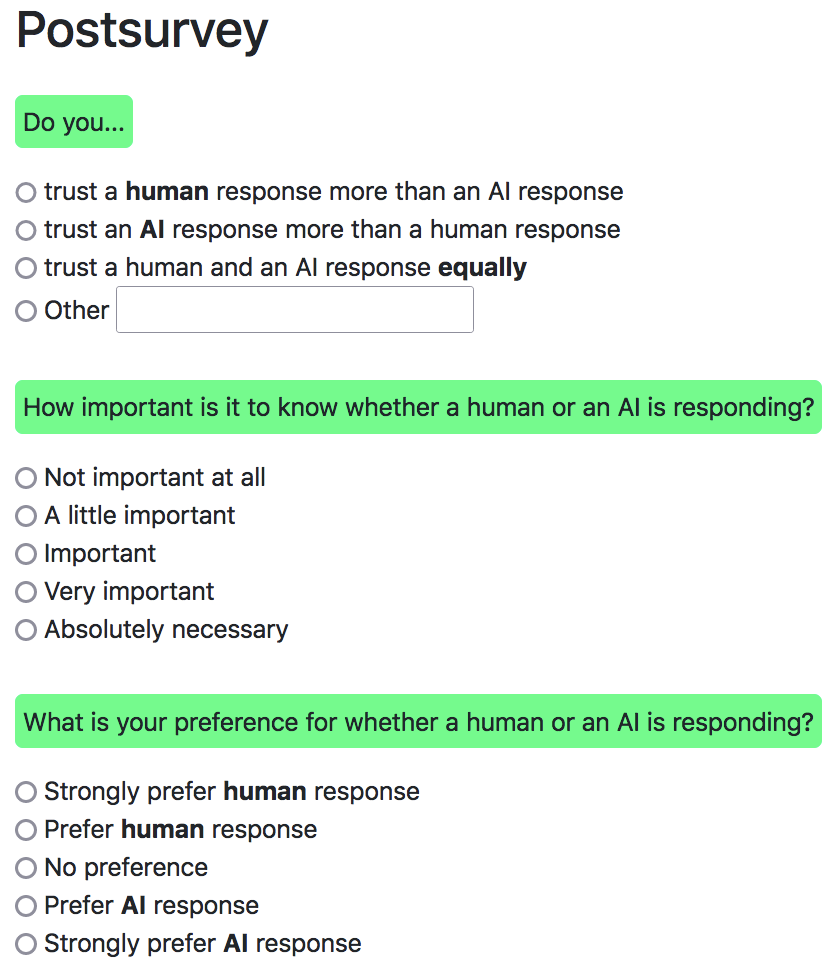}
    \label{fig:post1}
\end{figure}
\begin{figure}[!htb]
    \centering
    \includegraphics[width=.45\textwidth]{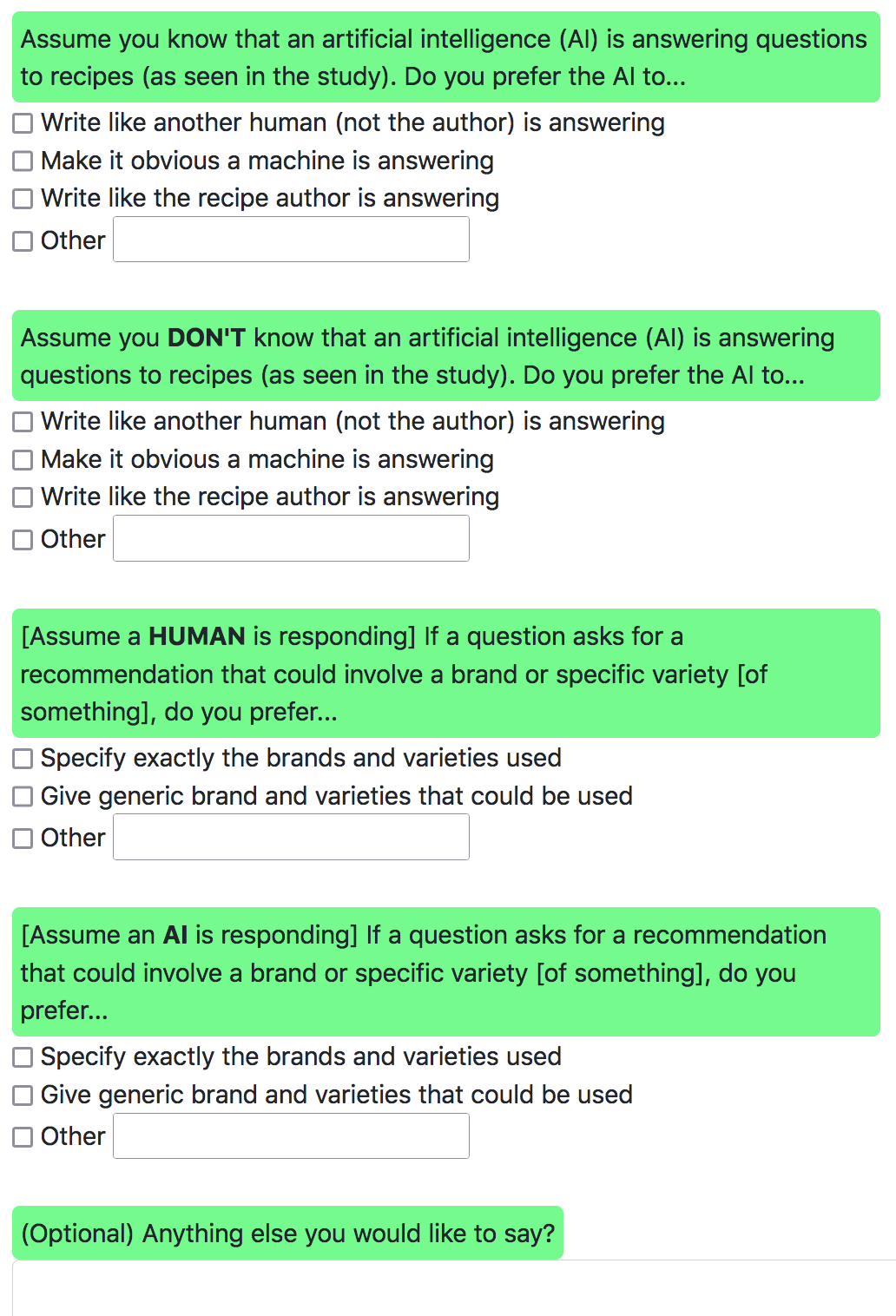}
    \caption{Postsurvey.}
    \label{fig:post2}
\end{figure}

\label{app:survey-results}
\subsection{Survey Results}
We summarize survey results of annotators where conclusions were drawn. 

\paragraph{Crowdworkers}
Six annotators had 7+ years of home baking and cooking experience, and all but one rated their cooking expertise as intermediate. Baking expertise had a larger range from beginner (3) to intermediate (5). Two annotators requested generic brands when a known machine is responding. Three annotators want a machine to make it obvious that a machine is responding regardless of whether it is known that a machine is responding. There was an even split in trusting human and machine responses equally or trusting a human more.

\paragraph{Experts}

Two experts had 1--3 years of professional baking and cooking experience, and one expert had 3--7 years of professional cooking experience. Baking experience was rated intermediate by all three, two experts stated they had intermediate-advanced cooking expertise, and one considered themself an expert. One expert wanted generic brands from a known machine, and one expert did not mind having specific or generic brands. Only one expert wanted a machine to make it obvious that a machine is responding when the respondent is unknown, and and another expert wanted similarly when it is known that a machine is responding. Two experts state they trust human responses over an AI response, and one trusts both human and AI responses equally.

\subsection{Improvement Results} \label{app:improvementResults}
Experts were more critical than crowdworkers for how responses could be improved and provided many custom suggestions for how to improve responses (Table \ref{tab:improvementResponses}).

\begin{table*}[hbt]
\centering
\resizebox{\linewidth}{!}{
\begin{tabular}{llrllrllr} 
\toprule
Area & AMT-GPT-3             & \%  & AMT-Human          & \%   & Expert-GPT-3       & \%   & Expert-Human          & \%    \\ 
\hline
Concise              & Ingr. quantity & 4.6 & Cook temp/time  & 2.4  & Cook temp/time  & 7.3  & Ingr. quantity & 6.9   \\
                     & Tech. Detail     & 1.8 & Tech. Detail  & 2.8  & Tech. Detail  & 18.3 & Tech. Detail     & 13.8  \\
                     & Expected results      & 8.5 & Expected results   & 2.9  & Expected results   & 15.6 & Expected results      & 8.6   \\
                     & Other                 & 6.8 & Other              & 10.9 & Other              & 19.2 & Other                 & 17.2  \\ 
\hline
Verbose              & Expected results      & 1.9 & Expected results   & 1.5  & Hallucination    & 4.2  & Hallucination       & 4.3   \\
                     & Other                 & 3.2 & Other              & 4.9  & Other              & 5.3  & Other                 & ---   \\ 
\hline
Misc                 & Hedging    & 2.8 & Hedging & 5.4  & Hedging & 15.2 & Hedging    & 2.1   \\
                     & Other                 & --- & Other              & ---  & Other              & 10.1 & Other                 & 6.4   \\
\bottomrule
\end{tabular}
}
\caption{Annotations for how responses could be improved.}\label{tab:improvementResponses}
\end{table*}

\begin{figure}
    \centering
    \includegraphics[width=.43\textwidth]{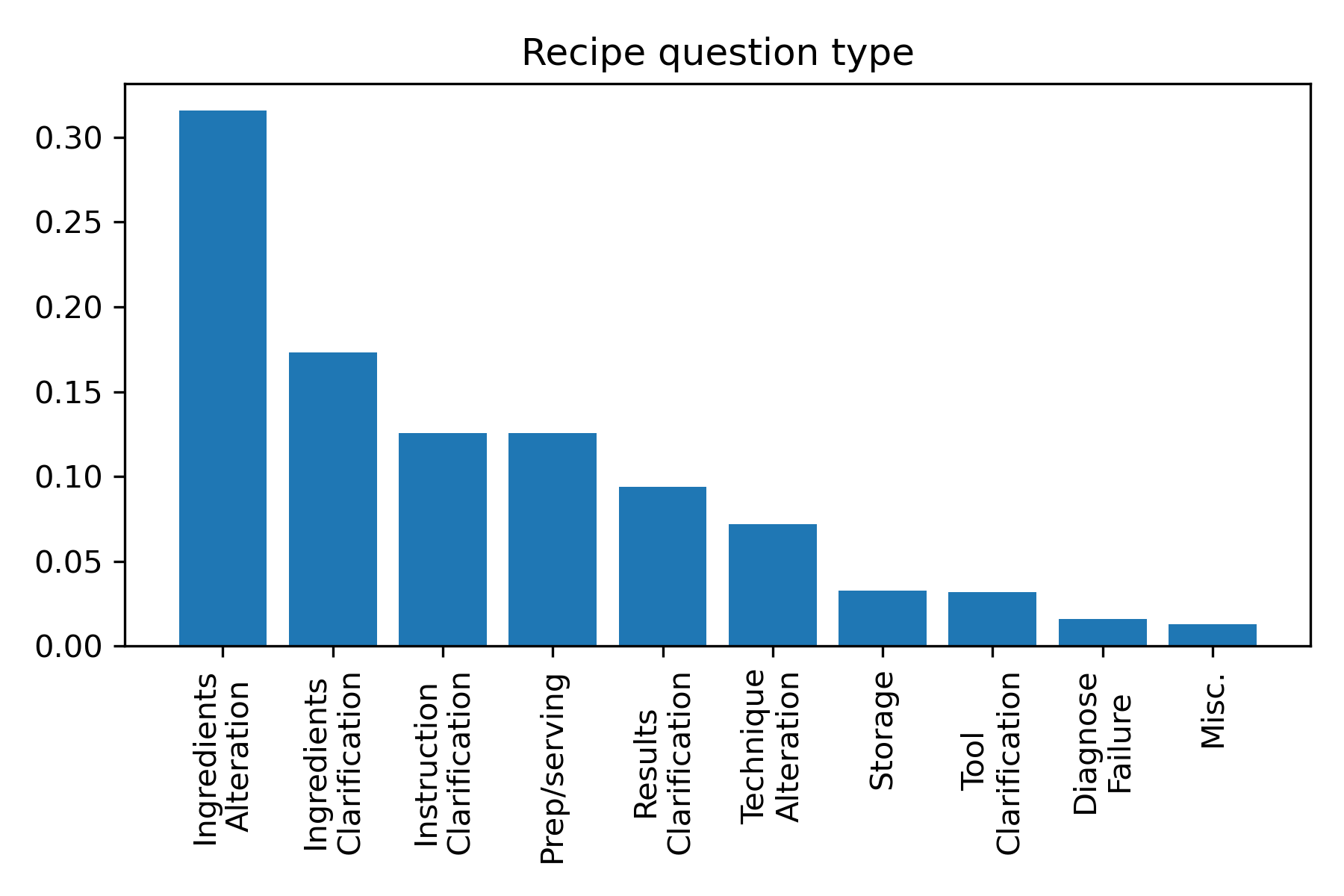}
    \caption{Distribution of the types of questions in the 60 questions annotated by both crowdworkers and experts. Misc. contains many infrequent custom question types. Storage was a frequently used custom type.}
    \label{fig:questionType}
\end{figure}

\subsection{Example Annotation Responses}\label{sec:exampleAnnotations}

\label{app:example-responses-annot-q}
\begin{figure*}
    \centering
    \includegraphics[width=.95\textwidth]{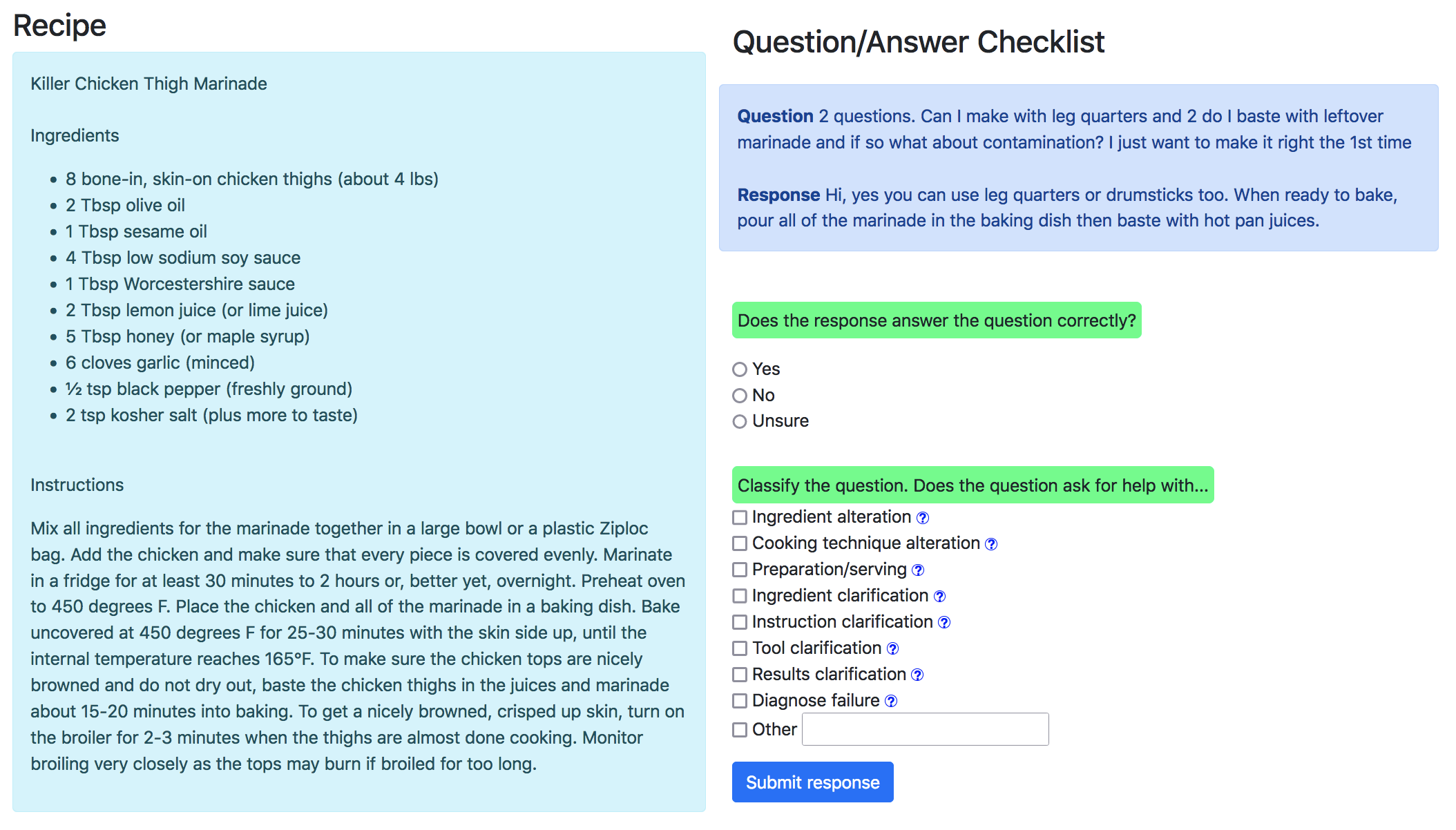}
    \caption{Example annotation question with responses described in \ref{app:example-responses-annot-q}.}
    \label{fig:example-annot-q}
\end{figure*}

Sample responses to a question in the annotation task (Figure \ref{fig:example-annot-q}) are as follows:
\begin{itemize}
    \item Yes, the response answers the question correctly. The question type was for ingredient alteration and clarification. Recipe context was not required to judge the correctness of the answer. The answer could be improved in the ``concise'' dimension by including more Tech. Detail and information on why the technique is safe. The satisfaction with the answer is 4 out of 5.
    \item No, the response does not answer the question correctly. The question type is cooking technique alteration and preparation/serving. Recipe context was not required to judge the correctness of the answer. The answer was incorrect because it described incorrect or unsafe processes.
\end{itemize}

\end{document}